\documentclass[final,3p,times]{elsarticle}

\usepackage[
monochrome
]{xcolor}

\usepackage{tikz}
\usetikzlibrary{trees,positioning,shapes,shadows,arrows}
\usepackage{forest}

\usetikzlibrary{mindmap}
\usepackage{amssymb}
\usepackage{comment}
\usepackage{mathtools}

\DeclarePairedDelimiter\floor{\lfloor}{\rfloor}
\usepackage{xcolor}
\usepackage{mathtools}
\usepackage{amsmath}

\usepackage{arydshln}
\usepackage{booktabs}
\usepackage{rotating}
\DeclareUnicodeCharacter{2061}{}
\usepackage[utf8]{inputenc}
\usepackage[T1]{fontenc}

\journal{Neurocomputing}

\begin{document}

\begin{frontmatter}



\title{A Gentle Introduction and Survey on \\ Computing with Words (CWW) Methodologies}


\author[l1]{Prashant~K.~Gupta}
\author[l2,l3]{Javier~Andreu-Perez}

\affiliation[l1]{organization={German Research Center for Artificial Intelligence (DFKI) GmbH},
            addressline={D3.1, Campus D3.2}, 
            city={Saarbrucken},
            postcode={66123}, 
            state={Saarland},
            country={Germany}}
            
\affiliation[l2]{organization={Centre for Computational Intelligence, School of Computer Science and Electronic Engineering, University of Essex},
            city={Colchester},
            country={United Kingdom}}

\affiliation[l3]{organization={Simbad2, University of Jaén},
            city={Jaén},
            country={Spain}}

\begin{abstract}
Human beings have an inherent capability to use linguistic information (LI) seamlessly even though it is vague and imprecise. Computing with Words (CWW) was proposed to impart computing systems with this capability of human beings. The interest in the field of CWW is evident from a number of publications on various CWW methodologies. These methodologies use different ways to model the semantics of the LI. However, to the best of our knowledge, the literature on these methodologies is mostly scattered and does not give an interested researcher a comprehensive but gentle guide about the notion and utility of these methodologies. Hence, to introduce the foundations and state-of-the-art CWW methodologies, we provide a concise but a wide-ranging coverage of them in a simple and easy to understand manner. We feel that the simplicity with which we give a high-quality review and introduction to the CWW methodologies is very useful for investigators, especially those embarking on the use of CWW for the first time. We also provide future research directions to build upon for the interested and motivated researchers.
\end{abstract}


\end{frontmatter}


\section{Introduction}\label{sec:introduction}
Computing with Words (CWW)\footnote{For readers' convenience, all the abbreviations are listed in Table \ref{tab:abbreviations}} was conceptualized and put forth in the research community for the first time by Prof. Zadeh\cite{zadeh1996}. The motivation behind the CWW was that if computing systems were built on the principles of CWW, they could think, reason, make decisions and solve problems using a ``concepts" driven analytical framework that resembles human symbolic cognition. Human cognition has a remarkable capability to process and reason using semantic uncertainty, which is inevitable in a number of day to day life situations. These situations require decision making using the variables which tend to take qualitative values, naturally and hence the information pertaining to the situation is vague, imprecise and semantically uncertain\footnote{Semantic uncertainty arises due to subjectivity.}. An example of such a situation can be deciding whom to befriend when meeting people at a gathering. Here, the decision is based on the perception of the nature or behavior of a person. The variable of interest viz., nature or behavior of a person, used here naturally tends to assume qualitative description and is thus vague, imprecise and contains semantic uncertainty. Another scenario can be the vagueness or imprecision in the semantic understanding of the knowledge provided by experts or decision-makers. 

A common observation is that the classical theory tends to use probabilistic treatment of qualitative concepts and thus tries to quantify the semantic uncertainty by allocating a precise number. Such concepts cannot be defined by boxing them on the basis of their odds of occurring. More often than not, in such situations, the uncertainty is of non-probabilistic nature based on its degree of truth. Thus, there was a need to search for soft information structures to describe these scenarios. This suitable information representation structure was the use of linguistic descriptors for the scenario. 

\begin{table}[h]
    \centering
    \caption{Abbreviations and Their Full Forms}
    \begin{tabular}{l l}
    \hline
    \hline
    Abbreviation &  Full-Form\\
    \hline
    AEPCM & Augmented Extension Principle based CWW methodology \\
    CWW & Computing with Words \\
    EPCM & Extension Principle based CWW methodology \\
    FOU & Footprint of Uncertainty \\
    FS & Fuzzy Sets \\
    GT2 & General Type-2 \\
    GFSCM & General Type-2 Fuzzy Sets based CWW methodology \\
    IFS & Intuitionistic Fuzzy Sets\\
    IFSCM & Intuitionistic Fuzzy Sets based CWW methodology \\
    IT2 & Interval Type-2 \\
    LGT2 & Linear General Type-2 \\
    LFSCM & Linear General Type-2 Fuzzy Sets based CWW methodology \\
    LMF & Lower Membership Function \\
    MF & Membership Fucntion \\
    RSCM & Rough Sets based CWW methodology \\
    SMCM & Symbolic Method based CWW methodology \\
    T1 & Type-1 \\
    2TPCM & 2-tuple based CWW methodology \\
    UMF & Upper Membership Function \\
    \hline
    \hline
    \end{tabular}
    \label{tab:abbreviations}
\end{table}

\begin{figure}[h]
    \centering
    \includegraphics[height=\textheight, width=\textwidth, keepaspectratio=true]{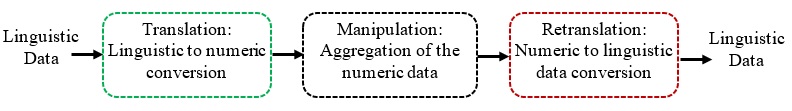}
    \caption{Prof. Yager's CWW Model \cite{yager1999computing}}
    \label{fig:yagergen}
\end{figure}

Human beings have a highly superior capability (with respect to animals) to effortlessly understand, process and operate using linguistic descriptors or `words'. Words make up the sentences in the natural language\footnote{By natural language is meant the language used by human beings for communication in day to day life.}. They are vague because people have different understanding and interpretations of the same `word'\cite{mendel2010perceptual}\footnote{According to Prof. Mendel, this can be specified in a sentence `words mean different things to different people'.}. Hence, CWW uses words as the units of computation. 

Over the years, massive literature has been proposed which presents the views of researchers on the CWW. Prof. Zadeh's other remarkable work in the CWW is \cite{zadeh1999computing}. Here, the importance of linguistic information over precise numeric measurements has been advocated by drawing a comparison between perception based information and the numeric measurements. The former is quite closely related to everyday human capabilities to do a number of tasks seamlessly in day to day life. Prof. Yager was of the opinion that Prof. Zadeh's CWW had a very specific way of using the words and hence gave a CWW Model \cite{yager1999computing}, which is made up of three steps (Please see Fig. \ref{fig:yagergen}). In the first step, called translation, a mapping or conversion of the linguistic information to its numeric counterpart is performed. The importance of this step is that a computing machine understands only numbers and hence cannot directly operate on the linguistic data. In the next step viz., manipulation, the mapped numeric information is aggregated. This emphasises that numeric information may come from various sources, or one source may provide numeric information multiple times, and thus aggregation is important for decision making. Finally, in the last step, called retranslation, the aggregated numeric information is mapped to linguistic form, as human beings attach comparatively more relevance to the linguistic information.

\begin{figure}[!h]
\resizebox{\textwidth}{!}{
\begin{tikzpicture}[mindmap, concept color = red!40,  every node/.style=concept, grow cyclic, text width=2.7cm, align=flush center]
    \node {CWW Methodologies}
        child [concept color=blue!30]{ node {T1 FSs}
            	child { node {EPCM}}
            	child { node {AEPCM}}
            	child { node {IFSCM}}
            }
        child [concept color=green!30] { node {Ordinal Term Sets}
            	child { node {SMCM}}
            	child { node {RSCM}}
            }
        child [concept color=yellow!30] { node {T1 FSs \& Ordinal Term Sets}
	            child { node {2TPCM}}
	        }
        child [concept color=pink!30]{ node {IT2 FSs}
	            child { node {Perceptual Computing}}
	       }
	       child [concept color=gray!30] {node {T1 FS extensions}
	            child { node {LFSCM}}
	            child { node {GFSCM}}
	        };
\end{tikzpicture}}
\caption{Mindmap of CWW Methodologies}
\label{fig:mindmap}
\end{figure}
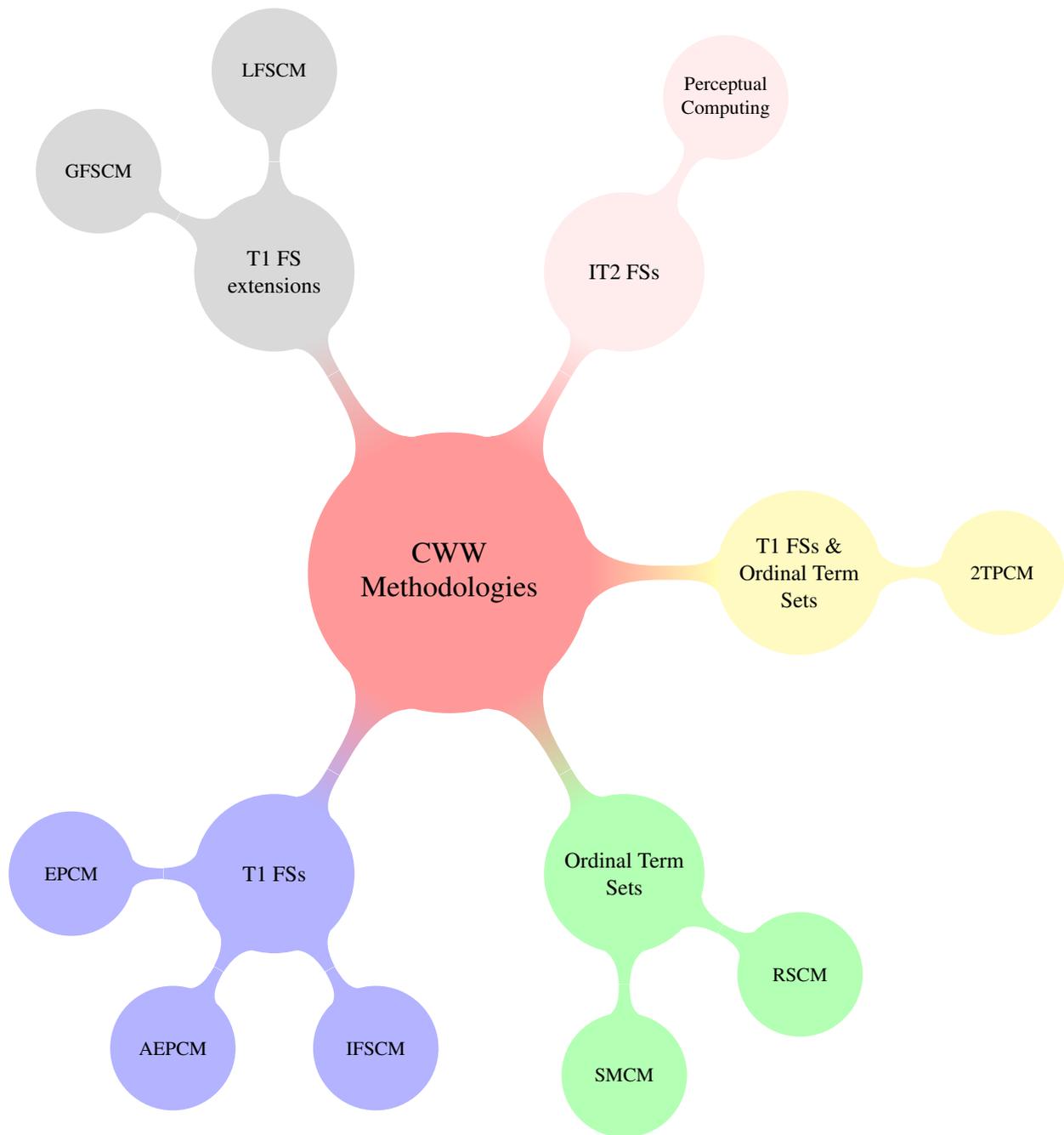

Recently there has been a surge in the area of CWW \textcolor{blue}{methodologies}, which is evident from \textcolor{blue}{numerous literary works being contributed in this direction.} These include mainly: Extension Principle based CWW methodology (EPCM) \cite{gupta2021enhanced}, Augmented Extension Principle based CWW methodology (AEPCM) \cite{gupta2021enhanced}, Intuitionistic Fuzzy Sets (IFS) based CWW methodology (IFSCM) \cite{gupta2021enhanced}, Symbolic Method based CWW methodology (SMCM) \cite{gupta2021enhanced}, Rough Sets based CWW methodology (RSCM) \cite{gupta2021enhanced}, 2-tuple based CWW methodology (2TPCM) \cite{herrera20002, martinez20152}, Perceptual Computing \cite{mendel2010perceptual}, Linear General Type-2 (LGT2) Fuzzy Sets based CWW methodology (LFSCM) \cite{bilgin2015ambient}, General Type-2 (GT2) Fuzzy Sets based CWW methodology (GFSCM\footnote{It is mentioned here that in work \cite{gupta2021enhanced}, the EPCM, AEPCM, IFSCM, SMCM and RSCM have been shown to process linguistic information just like the Yager's CWW model of Fig. \ref{fig:yagergen}. 2TPCM, Perceptual Computing, LFSCM and GFSM call these respective steps by different names.}) \cite{jiang2018general}.  These CWW methodologies use different units of uncertainty for modeling the semantics of the linguistic terms. The EPCM, AEPCM and IFSCM, use the type-1 (T1) fuzzy sets (FSs) to model the word semantics. The SMCM and RSCM, use the ordinal term sets. The 2TPCM performs information representation and computation using a combination of T1 FSs and ordinal term sets. The Perceptual Computing makes use of interval type-2 (IT2) FSs. The LFSCM and GFSCM use the GT2 FSs; however, in the former, the secondary membership function (MF) is a linear function, whereas it can be any arbitrary function in the latter.  \textcolor{blue}{A categorisation of these CWW methodologies is shown in the form of a mindmap in the Fig. \ref{fig:mindmap}.}

\textcolor{blue}{Table \ref{tab:comparisoncwwmethods} depicts the comparison of these CWW methodologies based on various criteria. It can be seen from the table that the criteria chosen for differentiating the CWW methodologies are primarily the instrument used to model the semantics of linguistic terms (LTs) and linguistic weights (LWs). The instrument is T1 FSs (for the EPCM, AEPCM and IFSCM), ordinal term sets (for SMCM and RSCM), a combination of T1 FSs as well as ordinal set (for 2TPCM), IT2 FSs (for perceptual computing) and GT2 FSs (for LFSCM and GFSCM). Further, to differentiate the CWW methodologies from each other within the respective instrument, a division is extended on the basis of how respective CWW methodology achieves CWW in respective three steps: translation, manipulation and retranslation of Yager's CWW (please see Fig. \ref{fig:yagergen}). For example, within T1 FSs based CWW methodology, there are three prominent placeholders: EPCM, AEPCM and IFSCM. EPCM converts T1 MFs of LTs in translation to tri-tuples. These are aggregated in manipulation and converted back to linguistic form using a linguistic approximation in the last step. AEPCM, on the other hand, approximate the T1 MFs of LTs and LWs using tri-tuples in translation, followed by weighted aggregation in manipulation and conversion back to linguistic form using a linguistic approximation in retranslation. IFSCM resorts to tri-tuple representation of membership and non-membership for both LTs and LWs in translation. This is followed by weighted aggregation and linguistic approximation in manipulation and retranslation, respectively. Further details about other CWW methodologies will be discussed in detail along with mathematical equations in Sections \ref{sec:l-1cwwmethodologiesbasedonordinal}-\ref{sec:l-4cwwmethodologies}.} 

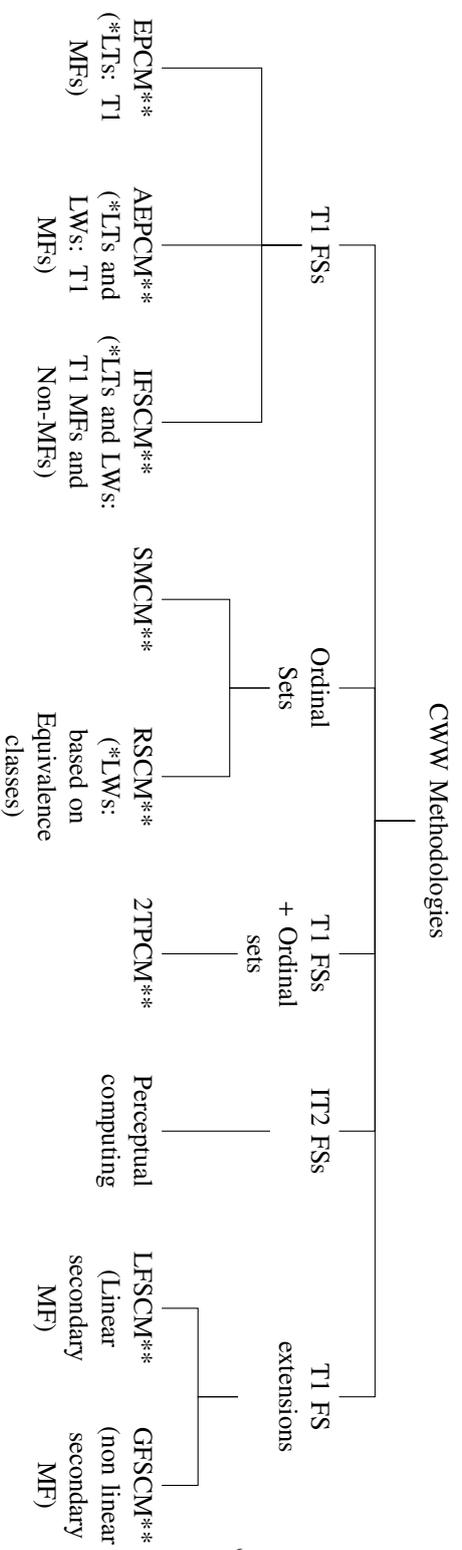
\begin{sidewaysfigure}
    \centering
\forestset{%
my forest/.style={%
    for tree={%
            node options={text width=5cm, align=center},
            l sep=1cm, 
            parent anchor=south, 
            child anchor=north,
            if n children=0{tier=word}{},
        edge path={%
            \noexpand\path [\forestoption{edge}] (!u.parent anchor) -- +(0,-15pt) -| (.child anchor)\forestoption{edge label};
        },
    }
}
}

\begin{forest} my forest,
[, for tree={s sep=-29mm} 
    [CWW Methodologies, no edge
        [T1 FSs, name=lvl1
           [EPCM** \linebreak (*LTs: T1 \linebreak MFs), name=lvl11]
           [AEPCM** \linebreak (*LTs and \linebreak LWs: T1 \linebreak MFs), name=lvl12]
           [IFSCM** \linebreak (*LTs and LWs: \linebreak T1 MFs and \linebreak Non-MFs), name=lvl13]
        ]
        [Ordinal \linebreak Sets, name=lvl2
           [SMCM**, name=lvl21]
           [RSCM** \linebreak (*LWs: \linebreak based on \linebreak Equivalence \linebreak classes), name=lvl22]
        ]
        [T1 FSs \linebreak + Ordinal \linebreak sets, name=lvl3
           [2TPCM**, name=lvl31]
        ]
        [IT2 FSs \linebreak, name=lvl4
           [Perceptual \linebreak computing, name=lvl41]
        ]
        [T1 FS \linebreak extensions \linebreak, name=lvl5
           [LFSCM** \linebreak (Linear \linebreak secondary \linebreak MF), name=lvl51]
           [GFSCM** \linebreak (non linear \linebreak secondary \linebreak MF), name=lvl52]
        ]
    ] 
]
\end{forest}
\caption{Taxonomy of CWW Approaches \\ *LTs= Linguistic terms, LWs= Linguistic Weights, **For full forms, please see Table \ref{tab:abbreviations}.}
\label{fig:taxonomy}
\end{sidewaysfigure}

\textcolor{blue}{An important fact needs mention here. We chose these criteria because it enables the reader to see the CWW methodologies in a holistic as well as detailed manner. To exemplify, one can see that using T1 FSs as the semantic modeling instrument, three CWW methodologies have been developed viz., EPCM, AEPCM and IFSCM, giving a holistic view of the T1 FSs based CWW methodologies. Each of these CWW methodologies has different internal working for processing the LI (please see Section \ref{sec:l-1cwwmethodologiesbasedonfss} for details), thus giving a detailed peek into the internal working of the T1 FSs based CWW methodologies. Similar argument applies to other CWW methodologies listed in Table \ref{tab:comparisoncwwmethods}. For the convenience of the readers, we have developed a taxonomy of these CWW methodologies, to highlight the differences between these methodologies. Please see Fig. \ref{fig:taxonomy}.}

\usetikzlibrary{positioning}

\setlength\dashlinedash{0.5pt}
\setlength\dashlinegap{1.5pt}
\setlength\arrayrulewidth{0.3pt}
\begin{sidewaystable}
\small
    \centering
    \caption{Comparison of CWW Methodologies}
    \begin{tabular}{c c c c c}
    \hline
    \hline
    CWW &  \multicolumn{4}{c}{Criteria}\\\cmidrule(lr){2-5}
    & MS** & LN** & A** & NL**\\
    Methodology* &\\
    \hline
    EPCM & LTs: using T1 MFs & Tri-tuples of T1 MFs & Arithmetic mean & Linguistic approximation\\
    \hdashline
    AEPCM & LTs and LWs: using T1 MFs & Tri-tuples of T1 MFs & Weighted Average & Linguistic approximation\\
    \hdashline
    IFSCM & Membership and non-membership & Tri-tuples of T1 MFs & Weighted Average & Linguistic approximation\\
    & of LTs and LWs: using T1 MFs & & &\\
    \hdashline
    SMCM & LTs: using ordinal term sets & Indices of ordinal terms & Recursion & Ordinal term set\\
    \hdashline
    RSCM & LTs: using ordinal term sets & Indices of ordinal terms & Recursion & Ordinal term set\\
    \hdashline
    2TPCM & LTs: using T1 MFs & Combination of T1 MFs & Weighted Average and & Symbolic translation \\
    & and Ordinal term sets & and ordinal term sets & Ordered Weighted Average & \\
    \hdashline
    Perceptual & IT2 FSs & IT2 FS word models & Interval weighted average, & Similarity, ranking,\\
    Computing & & & fuzzy weighted average and & and subsethood\\
    & & & linguistic weighted average & \\
    \hdashline
    LFSCM & T2 FSs; Secondary MF is a & Sensory mapping & Conversion to IF-THEN & Mapping and display \\
    & linear function for Shoulder FOUs & & rules and aggregation & to the user by GUI\\
    \hdashline
    GFSCM & Finite automata & State space & Transition function & Accepting state\\
    \hline
    \hline
    \multicolumn{5}{l}{*For full forms, please see Table \ref{tab:abbreviations}}\\
    \multicolumn{5}{l}{**MS= Methodology used for modelling semantics of linguistic terms (LTs) and/ or linguistic weights (LWs)}\\
    \multicolumn{5}{l}{**LN= Linguistic to numeric mapping, **A= Aggregation, **NL= Numeric to linguistic mapping}
    \end{tabular}
    \label{tab:comparisoncwwmethods}
\end{sidewaystable}

There have been various applications and theoretical research works on these CWW methodologies. EPCM has been used in clinical decision making \cite{degani1988problem} and multi-criteria decision making \cite{pedrycz2011fuzzy}. SMCM has been used for multi-person decision making \cite{yager1993non} and group decision making \cite{xu2004method}. An overview of the application of 2TPCM in decision making can be found in \cite{marti2012overview}, decision support system \cite{martinez2010computing}, risk assessment \cite{liu2010computing} and decision analysis \cite{malhotra2020systematic}. Theoretical works related to 2TPCM have been the proposition of a new 2-tuple model for CWW in \cite{wang2006new}, extended 2-tuple to deal with unbalanced term sets \cite{abchir2013towards}, etc. Perceptual computing has been applied for the design of social judgement advisor \cite{wu2010social}, hierarchical decision making \cite{wu2010computing}, investment judgement analysis \cite{mendel2010assisting}, journal publication decision making \cite{mendel2010journal}, love selection \cite{korjani2013love}, etc. Theoretical works on perceptual computing are \cite{liu2008encoding, wu2011enhanced, hao2015encoding, mendel2008perceptual, wu2007aggregation, rickard2011linguistic, karnik2001centroid, nie2014ensuring, mendel2016comparison, mendel2007computing, karnik2001operations}. The LFSCM has been applied for cooking recipe recommendations in Ambient Intelligent Environments \cite{bilgin2015ambient, bilgin2015linear, bilgin2014adaptive}. Other research works on the LFSCM are \cite{bilgin2013computing, bilgin2013towards, bilgin2013experience, bilgin2012general, bilgin2012towards}.  

\textcolor{blue}{These works provide some interesting conclusions. It can be seen that the literature on these CWW methodologies is scattered, which makes it difficult to grasp an idea of the field, especially for anyone starting the work on CWW methodologies. The works \cite{martinez20152} and \cite{mendel2018perceptual} tried to provide an overview of 2-tuple methodology and perceptual computer, respectively. They are a good start, however, they have quite limited focus. Hence, we found no single work which attempted to compile the literature on CWW methodologies and provide the interested readers with a holistic but easily comprehensible view. We feel strongly that the absence of such literature will hinder the interested readers to realize the utilities of these methodologies, understanding their subtle differences as well as developing a sense of their strengths-limitations.} 

\textcolor{blue}{All these have motivated us to put forth} a succinct but wide-ranging coverage of these methodologies, in a simple and easy to understand manner. We feel that the simplicity with which we give a high-quality review and introduction to the CWW methodologies, is very useful for investigators or especially for those embarking on the use of CWW for the first time. We also provide future research directions to build upon for the interested and motivated researchers.

\textcolor{blue}{The major contributions of this manuscript are as follows:
\begin{itemize}
    \item Providing a comprehensive guide and survey that introduces the most relevant state-of-the-art for CWW methodologies, which can always act as a manual for the interested readers.
    \item Serving as starting point reference to develop an understanding of the differentiating criteria, strengths and limitations of the CWW methodologies. 
    \item Putting forth substantial helpers and tables that compare the utility and properties of the CWW methodologies, in order to realize the potential of the field.
\end{itemize}
}

The remainder of this literary work follows the following organization: Section \ref{sec:l-1cwwmethodologiesbasedonfss} discusses EPCM, AEPCM and IFSCM; Section \ref{sec:l-1cwwmethodologiesbasedonordinal} gives details on SMCM and RSCM; Section \ref{sec:l-2cwwmethodologies} and Section \ref{sec:l-3cwwmethodologies} discusses the 2TPCM and Perceptual Computing, respectively. Section \ref{sec:l-4cwwmethodologies} gives the details of CWW methodologies based on T2 FSs or extensions of T1 FSs, Section \ref{sec:discussions} gives the important discussions, based on this research survey, and finally, Section \ref{sec:conclusions} concludes this article and puts forth its future scope. For the convenience of the text readers, we list down all the symbols used for describing the CWW methodologies in Table \ref{tab:symbols}.

\begin{table}[h]
    \centering
    \caption{Symbols used in CWW methodologies and Their meanings}
    \begin{tabular}{l l}
    \hline
    \hline
    Symbol &  Meaning\\
    \hline
    $T$ & Linguistic term set containing linguistic terms associated to \\
    & EPCM, AEPCM, IFSCM, SMCM\\
    $UP_{EP}$ & Collective preference vector containing feedbacks of the users for \\
    & EPCM, AEPCM, IFSCM, SMCM\\
    $UPT_{EP}$ & $UP_{EP}$ in tri-tuple form for EPCM\\
    $C$ & Collective Preference Vector obtained in Manipulation phase of \\
    & EPCM, AEPCM, IFSCM\\
    $d(t_q,C)$ & Distance between linguistic term $t_q$ and the preference vector $C$\\
    $W$ & Weight Vector Corresponding to weights of linguistic terms for \\
    & AEPCM, IFSCM, SMCM, RSCM \\
    $UPT_{A}$ & $UP_{EP}$ in tri-tuple form for AEPCM\\
    $UWT_{A}$ & $W$ in tri-tuple form for AEPCM\\
    $L_k \otimes W_k$ & Product of $k^{th}$ linguistic preference and its associated weight in AEPCM \\
    $UPT_{IFS}$ & $UP_{EP}$ in tri-tuple form for IFSCM\\
    $UWT_{IFS}$ & $W$ in tri-tuple form for IFSCM\\
    $d'(t_q,C)$ & Distance between linguistic term $t_q$ and the preference vector $C$ for\\
    & non-membership in IFSCM\\
    $UPS_{SM}$ & $UP_{EP}$ in sorted from according to indices for SMCM\\
    $AG^i$ & Recursive function for combining terms from $UPS_{SM}$ and $W$ in SMCM \\
    $AG^2$ & Boundary condition of $AG^i$ \\
    $UPS_{RS}$ & $UP_{EP}$ in sorted from according to indices for RSCM\\
    $AG'^i$ & Recursive function for combining terms from $UPS_{RS}$ and $W$ in RSCM \\
    $AG'^2$ & Boundary condition of $AG'^i$ \\
    $UP_{2TP}$ & $UP_{EP}$ converted to 2-tuple form for 2TPCM\\
    $W_{2TP}$ & $W$ converted to 2-tuple form for 2TPCM\\
    $\beta_{2tp}$ & Aggregation result of $UP_{2TP}$ and $W_{2TP}$ in 2TPCM \\
    $\alpha_{2tp}$ & Symbolic translation corresponding to $\beta_{2tp}$ \\
    $\tilde{Y}_{LWA}$ & Aggregation result obtained in Perceptual Computing \\
    $C_{\tilde{A}(x)}$ & Decoded centroid in Perceptual Computing \\
    $Q$ & Set of states used in GFSCM \\
    $\sum'$ & Set of symbols used in GFSCM \\
    $\delta$ & Type-2 fuzzy transition function used in GFSCM \\
    $q_0$ & Initial state used in GFSCM \\
    $F$ & GT2 FS of final states used in GFSCM \\
    \hline
    \hline
    \end{tabular}
    \label{tab:symbols}
\end{table}

\section{CWW based on T1 FSs}\label{sec:l-1cwwmethodologiesbasedonfss}
This section discusses the details of the CWW methodologies that model the word semantics using the T1 FSs. These include the EPCM, AEPCM and IFSCM. These methodologies achieve CWW in the same three steps as in Yager's CWW Model (Please see Fig. \ref{fig:yagergen}).

\subsection{EPCM}\label{subsec:extensionprinciplebasedcwwmethodology}
Let's assume that users are required to provide their linguistic preferences in a decision making scenario. These preferences come from a linguistic term set containing $g+1$ distinct linguistic labels. Examples of such labels can be `very low', `low', `medium', etc. Mathematically, if we denote the term set as $T$ and the linguistic labels or terms as $t_0,t_1,…t_g$, then in the set notation form, the term set can be denoted as: 

\begin{equation}
    \label{eq:Ttermset}
    T = \{t_0, ..., t_g\}
\end{equation}

\begin{figure}[t]
    \centering
    \includegraphics[height=0.7\textheight, width=0.7\textwidth, keepaspectratio=true]{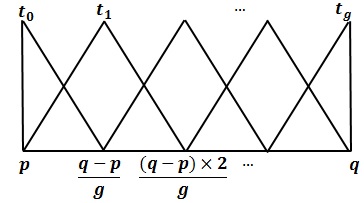}
    \caption{T1 FS based semantic representation of the terms in term set $T$ \cite{gupta2021enhanced}}
    \label{fig:T1-MFs-EP-SM}
\end{figure}

As stated earlier, the EPCM models the linguistic terms using the T1 FSs. However, these T1 FSs are generally in the form of uniformly shaped and distributed T1 triangular MFs on a bounded information representation scale. Let's assume an information representation scale whose ends are limited by $p$ and $q$\footnote{It is a common observation that frequently used values of $p$ and $q$, are respectively $0$ and $10$. However, any value can be used for them.}. Thus, the individual linguistic labels of the term set $T$ can be represented as shown in Fig. \ref{fig:T1-MFs-EP-SM}.

Let's say $i$ number of human subjects choose to elicit their preferences using the linguistic labels from the term set $T$ of (\ref{eq:Ttermset}). Hence, a set called the user preference vector, containing the user feedback can be given as:

\begin{equation}
    UP_{EP}=\{t_1,...t_i\}
    \label{eq:userprefs}
\end{equation}

where $UP_{EP}$ stands for user preferences in EPCM and and each $t_k \in T$; $k=1$ to $i$. This means that each $k^{th}$ index of the linguistic term in (\ref{eq:userprefs}) equals any one $j=0$ to $g$, in (\ref{eq:Ttermset}). The following subsections outline how EPCM processes these user feedbacks in three steps. 

\subsubsection{Step-1: Translation}
In the first step, the linguistic terms constituting the user preference vector, given in (\ref{eq:userprefs}), are mapped to numeric form. EPCM makes use of the T1 triangular MFs (for semantic representation of the linguistic terms) to perform this mapping, which is shown in Fig. \ref{fig:T1-MFs-EP-SM}. The numeric mappings of the linguistic labels given in (\ref{eq:userprefs}) are provided in the tri-tuple forms like $(l_k,m_k,r_k)$; $k=1$ to $i$. Here, $m_k$ corresponds to the midpoint of the triangular T1 MF where the MF achieves its highest value, whereas $l_k$ and $r_k$ correspond to the two ends of the T1 MF where it rests on the $x$-axis. The tri-tuples are shown in Fig. \ref{fig:T1-MFs-EP-SM}. Therefore, the modified user preference vector becomes:

\begin{equation}
    \label{eq:userpreftrituple}
    UPT_{EP}=\{(l_1,m_1,r_1),...(l_i,m_i,r_i)\}
\end{equation}

where $UPT_{EP}$ represents the set containing user preferences in tri-tuple form. 

\subsubsection{Step-2: Manipulation}
In the second step, the mapped numeric representations of the users preferences, given in (\ref{eq:userpreftrituple}), are aggregated. This is accomplished by averaging the respective $l_k$'s, $m_k$'s and $r_k$'s, to obtain the collective preference vector $C$ as:

\begin{equation}
    \begin{split}
        C&=(l_c,m_c,r_c)=(\frac{\sum_{k=1}^{i}l_k}{i},\frac{\sum_{k=1}^{i}m_k}{i},\frac{\sum_{k=1}^{i}r_k}{i}),\\
    l_c&=\frac{l_1+l_2+...+l_i}{i}, m_c=\frac{m_1+m_2+...+m_i}{i}, r_c=\frac{r_1+r_2+...+r_i}{i}
    \end{split}
    \label{eq:collectivepreferencevector}
\end{equation}

\subsubsection{Step-3: Retranslation}
In the final step, the numeric collective preference vector, given in (\ref{eq:collectivepreferencevector}), is mapped again to the linguistic form. This is useful for understanding by a human subject. As the linguistic terms, corresponding to the decision making scenario at hand, are contained in the linguistic term in $T$ of (\ref{eq:Ttermset}), therefore, the linguistic output from this step should also be a linguistic label from this term set. 

In order to map the collective preference vector of (\ref{eq:collectivepreferencevector}) to a linguistic label given in $T$ of (\ref{eq:Ttermset}), we calculate the Euclidean distance of the respective three defining points viz., $l_c,m_c,r_c$ of the preference vector from those of each linguistic term in $T$ of (\ref{eq:Ttermset})\footnote{Euclidean distance is used as a measure of similarity in \cite{gupta2021enhanced, herrera20002}. However, one is free to choose other measures such as support or cardinality.}. The computations are shown in (\ref{eq:eucledian}): 

\begin{equation}
    \label{eq:eucledian}
    d(t_q,C)=\sqrt{P_1(l_j-l_c)^2+P_2(m_j-m_c)^2+P_3(r_j-r_c)^2}
\end{equation}

where $t_j=(l_j,m_j,r_j )$, $j=0$ to $g$, is the tri-tuple representation of each linguistic terms in $T$ of (\ref{eq:Ttermset}), $C=(l_c,m_c,r_c)$, is the collective preference vector of (\ref{eq:collectivepreferencevector}), and $P_1 = P_3 = 0.2$ and $P_2 = 0.6$ are the weights. These $P_i,i=1,2,3$ values are taken from \cite{herrera20002}, however, no restriction exists on the use of a different set of values for respective $P_i$'s. Thus, the desired linguistic term, $t_b^* \in T$, is recommended on the basis of maximum similarity which is essentially equal to minimum Euclidean distance, which can be formally stated as $d(t_b^*,C) \leq d(t_j,C), \forall t_j \in T$.

\subsection{AEPCM}\label{subsec:augmentedextensionprinciplebasedcwwmethodology}
AEPCM was proposed in \cite{gupta2021enhanced}, as an improvement of EPCM (discussed in Section \ref{subsec:extensionprinciplebasedcwwmethodology}). The AEPCM can be utilized for achieving CWW in scenarios the users' linguistic preferences carry different importance and hence can be assigned different weights. The next subsections discuss the working of AEPCM. 

\subsubsection{Step-1: Translation}
Translation involves mapping the linguistic information to numeric form. Consider again the vector holding the containing $i$ users' linguistic feedback as given in (\ref{eq:userprefs}). Let's say that each of these users' feedback is assigned a linguistic weight, which can be similar or distinct from other weights. Hence, the vector containing the respective weights of the linguistic user preferences can be written as:

\begin{equation}
    \label{eq:weightvector}
    W=\{w_1,…,w_i\}
\end{equation}

where $w_p, p=1,...,i$ is the associated linguistic weight of the $p^{th}$ user's linguistic preference. It is mentioned here that the AEPCM chooses to model the semantics of each $w_p, p=1,...,i$ using uniformly shaped triangular T1 FS (similar to Fig. \ref{fig:T1-MFs-EP-SM}). Now, just like EPCM, the numeric mapping of each of these linguistic weights is a tri-tuple $\{l,m,r\}$, where $m$ corresponds to the MF point of highest membership values, whereas $l$ and $r$ are the points where the membership value is zero and triangular T1 MF touches the $x$-axis (please see (\ref{eq:userpreftrituple})). Thus, the users' preference vector (given in (\ref{eq:userprefs})), and the associated linguistic weights (given in (\ref{eq:weightvector})), when represented in the tri-tuple form are given respectively in (\ref{eq:usersprefstrituples-augeplcm}) and (\ref{eq:usersprefsweightstrituples-augeplcm}): 

\begin{equation}
    UPT_A=\{(l_1,m_1,r_1),...(l_i,m_i,r_i)\}
    \label{eq:usersprefstrituples-augeplcm}
\end{equation}

\begin{equation}
    UWT_A=\{(l_{w_1},m_{w_1},r_{w_1}),...(l_{w_i},m_{w_i},r_{w_i})\}
    \label{eq:usersprefsweightstrituples-augeplcm}
\end{equation}

where $UPT_A$ and $UWT_A$ stand for user preferences and associated weights, respectively in tri-tuple form for AEPCM. Each $(l_k,m_k,r_k)$,$k=1,…,i$ and $(l_{w_k},m_{w_k},r_{w_k})$, $k=1,…,i$ are three defining points of the triangular T1 MF for the user preferences and the respective linguistic weights.

\subsubsection{Step-2: Manipulation}
In this step, the mapped linguistic information from previous step is combined by performing the weighted aggregation, using the concept of $\alpha$-cuts given in \cite{gupta2021enhanced}. If the tri-tuple of a randomly selected linguistic user preference from (\ref{eq:usersprefstrituples-augeplcm}) and its associated weight from (\ref{eq:usersprefsweightstrituples-augeplcm}) are given respectively as $L_k=(l_k,m_k,r_k)$, $k=1,…,i$ and $W_k=(l_{w_k},m_{w_k},r_{w_k})$, $k=1,…,i$, then their product is given as: 

\begin{equation}
    \label{eq:produserprefweight-augeplcm}
        L_k \otimes W_k= \{ll_k,mm_k,rr_k\}= \{l_k \times l_{w_k}, m_k \times m_{w_k}, r_k \times r_{w_k}\}, k=1,2,...,i
\end{equation}

These products are obtained for each linguistic user preference as its associated weight. Hence, the collective preference vector, as an obtained by averaging the respective $ll_k$'s,$mm_k$'s, and $rr_k$'s is given as:

\begin{equation}
        C=(l_c,m_c,r_c)=(\frac{\sum_{k=1}^{i}ll_k}{i},\frac{\sum_{k=1}^{i}mm_k}{i},\frac{\sum_{k=1}^{i}rr_k}{i})
    \label{eq:collectivepreferencevector-augeplcm}
\end{equation}

\subsubsection{Step-3: Retranslation}
In the final step, we map the numeric collective preference vector of (\ref{eq:collectivepreferencevector-augeplcm}) to linguistic form, for the usefulness of end user. We perform the Euclidean distance based similarity computation as that of EPCM.

\subsection{IFSCM}\label{subsec:intuitionisticfuzzysetsbasedcwwmethodology}
In this section, we discuss the working of IFSCM. For basics on IFS, please see \ref{sec:basicsofifs}. 

\subsubsection{Step-1: Translation}
The first step viz., translation, will provide a linguistic to numeric mapping. Consider again the vector holding the linguistic feedbacks of $i$ number of users given in (\ref{eq:userprefs}) and the associated linguistic weights from (\ref{eq:weightvector}) of these linguistic feedbacks. As every element in an IFS has an associated degree of membership and non-membership (as shown in \ref{sec:basicsofifs}), therefore in IFSCM, each linguistic user preference and the associated weight is represented by a T1 MF. However, each preference and its respective weight has an associated membership and the non-membership degree. The membership degree value follows directly as the three defining points of the tri-tuples (as seen in Fig. \ref{fig:T1-MFs-EP-SM}). In contrast, the non-membership degree values are taken as the average of the tri-tuples of all linguistic terms except the one corresponding to the user's linguistic preference (or weight). 

For the purpose of elaboration, let's take up as an example the feedback of the first user from the term set of (\ref{eq:userprefs}) and its associated weight from (\ref{eq:weightvector}). The user's linguistic preference and its associated weight in the tri-tuple form can be written: $t_1=(l_1,m_1,r_1$) and $w_1=(l_{w_1},m_{w_1},r_{w_1})$\footnote{It's mentioned here that we have assumed that $t_1, w_1 \in T, T$ being given in (\ref{eq:Ttermset}). Therefore, $\exists t_p \in T,p=0,…,g,$ such that $t_1 = t_p$. Also $\exists t_q \in T, q=0,…,g,$ such that $w_1 = t_q$. Further, $t_p$ can be equal or unequal to $t_q$.}. The degree of non-membership for the user linguistic preference $t_1$ (respective linguistic weight $w_1$) is obtained by averaging respectively the three defining points $l_k,m_k,r_k$ of tri-tuples of all linguistic terms from $T$ except that of $t_p$ ($t_q$), which is mathematically given as:

\begin{equation}
    \label{eq:tri-tuplesnonmemtj1}
    (l'_1,m'_1,r'_1)=(\frac{\sum_{k=0, k \neq p}^{g}l_k}{g}, \frac{\sum_{k=0, k \neq p}^{g}m_k}{g},\frac{\sum_{k=0, k \neq p}^{g}r_k}{g})
\end{equation}

\begin{equation}
    \label{eq:tri-tuplesnonmemw1}
    (l'_{w_1},m'_{w_1},r'_{w_1})=(\frac{\sum_{k=0, k \neq p}^{g}l_k}{g}, \frac{\sum_{k=0, k \neq p}^{g}m_k}{g},\frac{\sum_{k=0, k \neq p}^{g}r_k}{g})
\end{equation}

In this manner, the membership degrees and non-membership degrees can be computed for all the linguistic preferences in (\ref{eq:userprefs}) and their associated weights in (\ref{eq:weightvector}). Hence, the vector depicting the collection of tri-tuples for the memberships degrees and non-membership degrees for linguistic preferences (associated weights) is given in (\ref{eq:usersprefstrituples-ifs}) ((\ref{eq:usersprefsweightstrituples-ifs})) as: 

\begin{equation}
    \label{eq:usersprefstrituples-ifs}
        UPT_{IFS}={[(l_1,m_1,r_1),(l'_1,m'_1,r'_1)],…,[(l_i,m_i,r_i),(l'_i,m'_i,r'_i)]}
\end{equation}

\begin{equation}
    \label{eq:usersprefsweightstrituples-ifs}
        UWT_{IFS}={[(l_{w_1},m_{w_1},r_{w_1}),(l'_{w_1},m'_{w_1},r'_{w_1})],…,[(l_{w_i},m_{w_i},r_{w_i}),(l'_{w_i},m'_{w_i},r'_{w_i})]}
\end{equation}

Here, $m_k (m_{w_k})$, $k=1,…,i$ is the point where the degree of membership for linguistic user preference (associated linguistic weight) achieves its maximum value, where as $l_k,r_k ((l_{w_k},r_{w_k}))$, $k=1,…,i$ are two points where the degree of membership for linguistic user preference (associated linguistic weight) has a null value and rests on $x$-axis. The $l'_k,m'_k,r'_k ((l'_{w_k},m'_{w_k},r'_{w_k}))$ are the corresponding values for the degree on non-membership for linguistic user preference (associated respective linguistic weight).

\subsubsection{Step-2: Manipulation}
The next step is to perform the aggregation of the weighted linguistic preferences and the respective associated linguistic weights. However, as each one is represented in the form of T1 MFs, hence we need to use the $\alpha$-cuts (similar to Section \ref{subsec:augmentedextensionprinciplebasedcwwmethodology}). Further, an added processing in IFS comes from performing these computations on the degree of memberships and non-memberships, separately for each user preference. Once this is done, next we aggregate the degrees of memberships (non-memberships) by averaging the three defining points of the weighted user preferences to get the collective preference vector as:

\begin{equation}
    \label{eq:preferencevector-ifs}
    \begin{split}
        &C=\{(l_c,m_c,r_c),(l'_c,m'_c,r'_c)\}\\
        &(l_c,m_c,r_c)=(\frac{\sum_{k=1}^{i}l_k}{i},\frac{\sum_{k=1}^{i}m_k}{i},\frac{\sum_{k=1}^{i}r_k}{i})\\
        &(l'_c,m'_c,r'_c)=(\frac{\sum_{k=1}^{i}l'_k}{i},\frac{\sum_{k=1}^{i}m'_k}{i},\frac{\sum_{k=1}^{i}r'_k}{i})
    \end{split}
\end{equation}

Here, $(l_c,m_c,r_c)$ corresponds to the membership and $(l'_c,m'_c,r'_c)$ corresponds to the non-membership. 

\subsubsection{Step-3: Retranslation}
Now we map the aggregated numeric data from the manipulation step into the linguistic form, using the Euclidean distance as a similarity measure as done for EPCM and AEPCM. The difference lies in the fact that the mapping is performed for each of the membership and non-membership values in the collective preference of (\ref{eq:preferencevector-ifs}), which is given as: 

\begin{equation}
    \label{eq:eucledian-ifs-mem}
    d(t_q,C)=\sqrt{P_1(l_j-l_c)^2+P_2(m_j-m_c)^2+P_3(r_j-r_c)^2}
\end{equation}

\begin{equation}
    \label{eq:eucledian-ifs-nonmem}
    d'(t_q,C)=\sqrt{P_1(l_j-l'_c)^2+P_2(m_j-m'_c)^2+P_3(r_j-r'_c)^2}
\end{equation}

where the $P_i,i=1,2,3$ are the weights, with values $0.2, 0.6$ and $0.2$ respectively. 

Finally, there are two recommended linguistic terms viz., $t^*_b \in T$ and $t^{*'}_b \in T$, corresponding respectively to the membership and non-membership.

\section{CWW based on Ordinal term sets}\label{sec:l-1cwwmethodologiesbasedonordinal}
In this section, we discuss the CWW methodologies which use the ordinal term sets for achieving the CWW. These are the SMCM and the RSCM. It is mentioned here that both SMCM and RSCM can process the differentially weighted linguistic user preferences. Also, both these methodologies achieve CWW in three steps viz., translation, manipulation and retranslation.

\subsection{SMCM}\label{subsec:symbolicmethodbasedcwwmethodology}
Let's now discuss the working of the SMCM, where the user preferences, as well as the associated respective weights, are represented through their indices in the term sets. 

\subsubsection{Step-1: Translation}
The starting point of the symbolic method based linguistic computational model is a linguistic preference vector that includes the user preferences. Consider the linguistic preference set from (\ref{eq:userprefs}). Each of these linguistic preferences may have an associated weight, given in the form of a weight vector given in (\ref{eq:weightvector}). An additional condition that SMCM imposes on this weight vector is that all the weights must add to $1$. This is stated as $\forall w_p \in W, w_p \in [0,1]; p=1$ to $i$, is the weight associated to the $p^{th}$ user linguistic preference in the term set given in (\ref{eq:userprefs}).

\subsubsection{Step-2: Manipulation}
In the second step, the linguistic user preferences from (\ref{eq:userprefs}) are sorted in descending order according to the indices of the linguistic terms drawn from $T$ of (\ref{eq:Ttermset}). After ordering, the user linguistic preferences may be given as:

\begin{equation}
    \label{eq:userprefssorted}
    UPS_{SM}=\{T_1,...T_i\}
\end{equation}

where $UPS_{SM}$ stands for user preferences in sorted order for the SMCM, $T_k \in T, k=1,…,i$. The linguistic preference vector from (\ref{eq:userprefssorted}) is then order weighted aggregated using the recursive function $(AG^i)$, given as:

\begin{equation}
    \label{eq:agrecursivemorethan2}
    \begin{split}
        &AG^i\{w_p, I_{T_p}, p=1,2,...,i \mid i > 2, i \in Z\}\\
        &= \{w_1 \odot I_{T_1}\} \oplus \{(1-w_1) \odot AG^{i-1}\{\delta_h, I_{T_h}, h=2,...,i\}\}
    \end{split}
\end{equation}

where $I_{T_p},p=1,…,i, I_{T_{jh}},h=2,…,i$ are the indices of the linguistic terms given in (\ref{eq:userprefssorted}) and $\delta_h=\frac{w_h}{\sum_{l=2}^{i}w_l};h=2,3,…,i$. As the aggregation function given in (\ref{eq:agrecursivemorethan2}), is a recursive function, therefore, we need a base condition when the recursion bottoms out. This is achieved, when the number of terms to be aggregated is reached at two. Hence, upon reaching the boundary condition, the aggregation function becomes as shown in (\ref{eq:agrecursiveequalto2}): 

\begin{equation}
    \label{eq:agrecursiveequalto2}
        AG^2\left\{\{w_{i-1}, w_i\}, \{I_{T_{i-1}}, I_{T_i}\},\mid i = 2\right\} = \{w_{i-1} \odot I_{T_{i-1}}\} \oplus \{w_i \odot I_{T_i}\}
\end{equation}

where $I_{T_{i-1}}$ and $I_{T_i}$ are the respective indices of the remaining terms from the preference vector (\ref{eq:userprefssorted}), with respective weights $w_{i-1}$ and $w_i$. Hence, the linguistic term is recommended in the next step.

\subsubsection{Step-3: Retranslation}
In this step, a linguistic term is recommended in the output. As a starting point for the same, for the (\ref{eq:agrecursiveequalto2}), a numeric index value is recommended using the computations shown in (\ref{eq:recommendedindex}) as:

\begin{equation}
    I_r=min \{i,I_{T_i}+round(\frac{w_{i-1}-w_i+1}{2}.(I_{T_{i-1}}-I_{T_i}))\}
    \label{eq:recommendedindex}
\end{equation}

here $round()$ is the round function, given as $round(x)= \floor*{x+0.5},x \in R, \floor*{}$ being the floor function\footnote{A special case of (\ref{eq:recommendedindex}) can arise when $w_{i-1}=w$ and $w_i=1-w$, for a random value of the weight $w$. Hence, rewriting the indices $I_{T_{i-1}}=I_l$ and $I_{T_i}=I_q$, (\ref{eq:agrecursiveequalto2}) and (\ref{eq:recommendedindex}) get modified to (\ref{eq:agrecursiveequalto2liq}) and (\ref{eq:recommendedindexiliq}), respectively as:

\begin{equation}
    \label{eq:agrecursiveequalto2liq}
    AG^2\left\{\{w, 1-w\}, \{I_l, I_q\}\right\}= \{w \odot I_l\} \oplus \{(1-w) \odot I_q\}
\end{equation}

\begin{equation}
    \label{eq:recommendedindexiliq}
    I_r=min\{i, I_q+round(w.(I_l-I_q))\}
\end{equation}}.

Now, starting from (\ref{eq:agrecursivemorethan2}), the recursive function $AG^{i}$ is called $AG^{i-1}$, which in turn calls for $AG^{i-2}$ and so on until $AG^2$ is reached, where the recursion bottoms out. The recommended numeric index $I_r$ using (\ref{eq:recommendedindex}) for $AG^2$ is then fed to $AG^3$, which again recommends a linguistic term. Thus, backtracking $i-2$ intermediate recursive equations in this manner, we reach the original recursive function $AG^{i}$ and recommend a numeric index for it too\footnote{For detailed computations, please see \cite{gupta2021enhanced}}. 

Finally, the recommended numeric index for the recursive function $AG^{i}$ can be matched to one of the terms from (\ref{eq:Ttermset}), to generate a linguistic recommendation in the output.

\subsection{RSCM}\label{subsec:roughsetsbasedcwwmethodology}
The RSCM is based on the rough sets \cite{dubois1990rough, lin1999granular, young2005granular, pawlak1982rough, pawlak1995rough}, which are obtained from crisp sets by drawing lower and upper approximations. The RSCM uses the indiscernibility property of rough sets and the concepts of SMCM. For details on indiscernibility, please see \ref{sec:indiscernibilityroughsets}.

\subsubsection{Step-1: Translation}
For the purpose of illustrating the working of linguistic to numeric mapping in RSCM, let's take the starting point as the linguistic feedbacks of $i$ users in the vector given in (\ref{eq:userprefs}). RSCM proceeds by dividing these user feedbacks into equivalence classes using the indiscernibility property. Hence, the vector containing users' feedbacks takes the form of a vector of equivalence classes and is given in (\ref{eq:equivalenceclasses-roughsets}) as:

\begin{equation}
    \label{eq:equivalenceclasses-roughsets}
    \{C_1,C_2,...C_n\}
\end{equation}

where each $C_i,i=1,…,n$ or an equivalence class is a set containing same linguistic preferences grouped together and defined as $C_i=(t_1,t_2,...t_p)$, where each $t_q \in T;q=1$ to $p,t_1= t_2=...=t_p$, $T$ being taken from (\ref{eq:Ttermset}). As each $C_i$ is a set, therefore we define the class cardinality or $\mid C_i \mid,i=1,…,n$, as the number of linguistic preferences constituting the class.

From Section \ref{subsec:symbolicmethodbasedcwwmethodology} it follows that the summation of the respectively associated weights of the corresponding linguistic preferences yields 1. As there are $n$ equivalence classes in (\ref{eq:equivalenceclasses-roughsets}), therefore, each class receives a weight of $1/n$. Further, since the class cardinality is $\mid C_i \mid,i=1,…,n$, so every linguistic term within a class can be allocated a weight of $1/(n \times \mid C_i \mid)$. Consequently, each linguistic preferences given in (\ref{eq:userprefs}) is allocated a weight, and the new weight vector can be given in (\ref{eq:weightvector-roughsets}) as:

\begin{equation}
    \label{eq:weightvector-roughsets}
    W=\{w'_1,…,w'_i\}
\end{equation}

here each $w'_p \in [0,1];p=1$ to $i$ is the respective associated weight of the user feedback taken from (\ref{eq:userprefs}). Also, it follows trivially that $\sum_{p=1}^{i}w'_p =1$.

\subsubsection{Step-2: Manipulation}
The aggregation in RSCM is performed similarly to that of SMCM discussed in Section \ref{subsec:symbolicmethodbasedcwwmethodology}, using the recursive function. Initially, the linguistic preferences are sorted according to their indices, and therefore the new vector depicting the user preferences becomes: 

\begin{equation}
    \label{eq:userprefsorted-roughsets}
    UPS_{RS} = \{T_1,T_2,...T_i\}
\end{equation}

where $T_k \in T,k=1,…,i$. It is mentioned here that each of the $T_k$ may or may not be equal to respective $t_k$. RSCM uses a recursive function $(AG'^i)$ for aggregation similar to SMCM and is given in (\ref{eq:recursivefunction-rough-morethan2}) as:  

\begin{equation}
    \label{eq:recursivefunction-rough-morethan2}
    \begin{split}
        &AG'^i\{w'_p, I_{T_p}, p=1,2,...,i \mid i > 2, i \in Z\}\\
        &= \{w'_1 \odot I_{T_1}\} \oplus \{(1-w'_1) \odot AG'^{i-1}\{\delta_h, I_{T_h}, h=2,...,i\}\}
    \end{split}
\end{equation}

where $I_{T_p},p=1,…,i, I_{T_h},h=2,…,i$ are the indices of the linguistic terms given in (\ref{eq:userprefsorted-roughsets}) and $\delta_h=\frac{w'_h}{\sum_{l=2}^{i}w'_l};h=2,3,…,i$.  
The recursive function $(AG'^i)$ reaches a boundary condition when the number of terms to be aggregated is two, when the recursive function looks like as shown in (\ref{eq:recursivefunction-rough-equal-to-2}): 

\begin{equation}
    \label{eq:recursivefunction-rough-equal-to-2}
        AG'^2\left\{\{w'_{i-1}, w'_i\}, \{I_{T_{i-1}}, I_{T_i}\},\mid i = 2\right\}= \{w'_{i-1} \odot I_{T_{i-1}}\} \oplus \{w'_i \odot I_{T_i}\}
\end{equation}

where $I_{T_{i-1}}$ and $I_{T_i}$ are the indices of the remaining terms from the user linguistic preference vector (\ref{eq:userprefsorted-roughsets}), and $w'_{i-1}$ and $w'_i$ are their respective associated weights.

\subsubsection{Step-3: Retranslation}
Now we generate a numeric index for the aggregated user preferences from the previous step. This numeric value is mapped into an output linguistic recommendation. The numeric index of the term for (\ref{eq:recursivefunction-rough-equal-to-2}) is $I_r$, given as:

\begin{equation}
    \label{eq:recommended-roughsets}
    I_r=min \left\{i,I_{T_i}+round\left(\frac{w'_{i-1}-w'_i+1}{2}.(I_{T_{i-1}}-I_{T_i})\right)\right\}
\end{equation}

where $round()$is the round function\footnote{A special case of (\ref{eq:recommended-roughsets}) is possible when $w'_{i-1}=w$ and $w'_i=1-w$, for any weight $w$. Hence, denoting $I_{T_{i-1}}=I_l$ and $I_{T_i}=I_q$, the boundary condition and the recommended term index equations can be given as:

\begin{equation}
    \label{eq:recursiveequalto2liq-roughsets}
    AG'^2\left\{\{w, 1-w\}, \{I_l, I_q\}\right\}= \{w \odot I_l\} \oplus \{(1-w) \odot I_q\}
\end{equation}

\begin{equation}
    \label{eq:recommendedindexiliq-roughsets}
    I_r=min\{i, I_q+round(w.(I_l-I_q))\}
\end{equation}}.

Thus, similar to SMCM, we started the aggregation process with the recursive function $AG'^i$ in (\ref{eq:recursivefunction-rough-morethan2}), which calls the function $AG'^{i-1}$, which will in turn call function $AG'^{i-2}$ and so on till we reach $AG'^3$ which in the boundary step case calls $AG'^2$ in (\ref{eq:recursivefunction-rough-equal-to-2}), thus in all we pass through $i-2$ intermediate recursion calls. The recommended index for $AG'^2$ is calculated using (\ref{eq:recommended-roughsets}). From here, we backtrack and will use this value as input to $AG'^3$. Hence, backtracking through a series of $i-2$ calls, the final recommended numeric index for $AG'^i$ can be found. This recommended index is matched to terms in (\ref{eq:Ttermset}) to generate a linguistic recommendation.

\section{CWW based on 2TPCM}\label{sec:l-2cwwmethodologies}
CWW can be achieved by a novel methodology called the 2TPCM, which uses a combination of T1 FSs and ordinal term sets. T1 FSs are used to model the semantics of the linguistic terms, whereas the processing involves using the indices of these terms in the term set. In the term set, these terms are represented using the ordinal term sets. The 2TPCM achieves CWW in three steps viz., information gathering phase, aggregation phase and exploitation phase.

\subsection{Step-1: Information gathering Phase}
In the 2TPCM, every piece of the linguistic preference is represented as twin valued viz., the linguistic term and its translation or distance from the nearest linguistic term (called symbolic translation) in the term set. Simply for the purpose of illustration, consider a term set as: $\{s_0: VL, s_1: L, s_2: M, s_3: H, s_4: VH\}$, where $VL$ stands for $Very~Low$, identified by the index $s_0$ or $0$. The $L$, $M$, $H$ and $VH$ stand respectively for $Low$, $Medium$, $High$ and $Very~High$, with respective indices being $s_1$ ($1$), $s_2$ ($2$), $s_3$ ($3$) and $s_4$ ($4$). Thus, a piece of numeric information, say $3.2$ is represented as $(s_3, +0.2)$ or $(H, +0.2)$ viz., it is at a distance of $+0.2$ from the linguistic term $H$. Further, a numeric information with a value $3.8$ is represented as $(s_4, -0.2)$ or $(VH, -0.2)$ viz., it is at a distance of $-0.2$ from the linguistic term $VH$.

Consider again the user preferences given in (\ref{eq:userprefs}) and their associated linguistic weights in (\ref{eq:weightvector}). The user preferences and the respective associated weights are represented as: $(t_k,\alpha), t_k \in T$, $T$ being the term set given in (\ref{eq:Ttermset}), and $\alpha$, called the symbolic translation is given as $\alpha \in [-0.5, 0.5)$. Thus, the two sets are given as:

\begin{equation}
    UP_{2TP}=\{(t_1, \alpha_1),...,(t_i, \alpha_i)\}
    \label{eq:userprefs2tp}
\end{equation}

\begin{equation}
    W_{2TP}=\{(w_1, \alpha_1),...,(w_i, \alpha_i)\}
    \label{eq:weightprefs2tp}
\end{equation}

As each of $t_k \in T$ and $w_k \in T$, $k=1$ to $i$, therefore all the $\alpha_k=0$. 

\subsection{Step-2: Aggregation Phase}
The users' feedbacks and their associated weights from (\ref{eq:userprefs2tp}) and (\ref{eq:weightprefs2tp}) respectively, are aggregated using the weighted arithmetic mean to compute the aggregation value $\beta_{2tp}$ as: 

\begin{equation}
    \label{eq:aggregation2tp}
    \beta_{2tp} = \frac{(w_1 \times I_{t_1})+...+(w_i \times I_{t_i})}{w_1+ ... +w_i}
\end{equation}
where $I_{t_k}$, $k = 1, ..., i$ are the indices of the linguistic terms are given in (\ref{eq:userprefs2tp}).

\subsection{Step-3: Exploitation Phase}
In this step, the $\beta_{2tp}$, obtained by aggregation in (\ref{eq:aggregation2tp}), needs to be converted back to the linguistic form. For this purpose, the symbolic translation $\alpha_{2tp}$ for the $\beta_{2tp}$ in (\ref{eq:aggregation2tp}) is given as:

\begin{equation}
    \label{eq:alpha2tp}
    \alpha_{2tp} = \beta_{2tp} - round(\beta_{2tp}), -0.5 \leq \alpha_{2tp} < 0.5
\end{equation}

Finally, the linguistic term output from the exploitation phase is: 

\begin{equation}
    \label{eq:recommend2tp}
    t_{recommended} = (t_{round(\beta_{2tp})}, \alpha_{2tp})
\end{equation}
The linguistic value of the recommended linguistic term is obtained by matching $t_{round(\beta_{2tp})}$ to the terms of $T$, of (\ref{eq:Ttermset}).

\section{CWW based on Perceptual Computing}\label{sec:l-3cwwmethodologies}
\begin{figure}[h]
    \centering
    \includegraphics[height=\textheight, width=\textwidth, keepaspectratio=true]{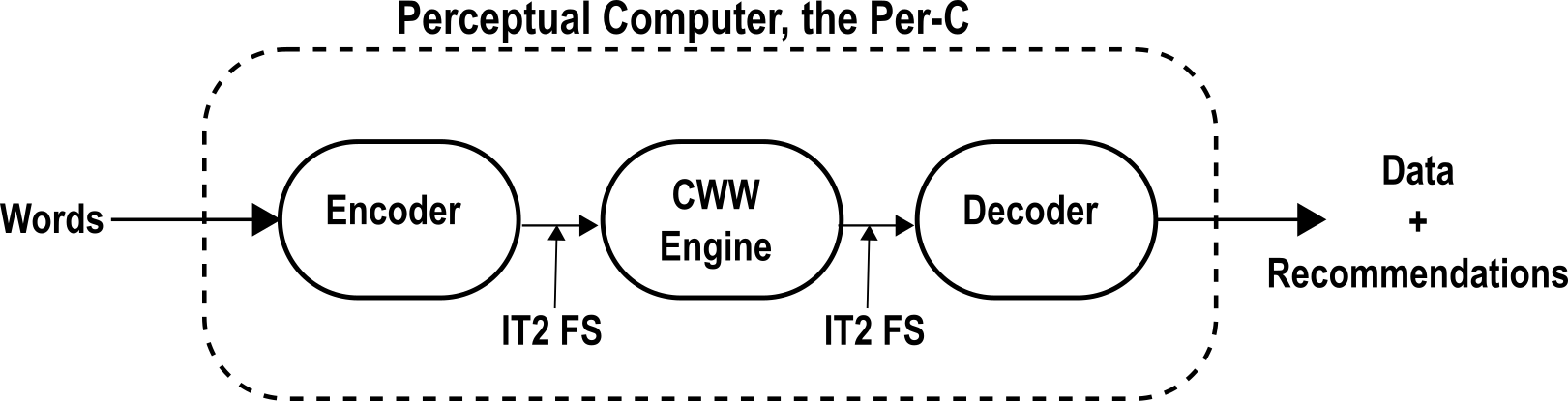}
    \caption{The Perceptual Computer (Per-C)}
    \label{fig:perc}
\end{figure}

Perceptual Computing is a novel CWW methodology that models the word semantics using IT2 FSs. Perceptual computing also achieves CWW in three steps similar to that of Yager's CWW framework, as shown in Fig. \ref{fig:yagergen}. However, the three steps are called encoder, CWW engine and decoder. The three steps are bound in a framework called the perceptual computer or Per-C, which is shown in Fig. \ref{fig:perc}. We now discuss the internal working of the Per-C in detail.

\subsection{Encoder}\label{subsec:percencoder}
The encoder converts the words into their nine point IT2 FS models, which form the numeric representations for the linguistic information. The words and their IT2 FS word models are stored together in a codebook. First of all, a vocabulary of problem-specific words is decided. Following this, the generating of the respective IT2 FS word model for a word involves the collection of endpoint data intervals from a group of subjects, generally through a survey. The subjects are asked to provide their opinions about the locations of endpoints. They are asked a question, “Assume that the endpoints of the word can be located on a scale of 0 to 10. Where do you think the endpoints of the word lie?”

Scenarios may arise where the subjects are unavailable, or the system needs to be designed to generate personalized recommendations. By personalized recommendations, we mean that the requirements of an individual user play a greater role in the system design than the group of users. For such cases, a novel approach of Person Footprint of Uncertainty (FOU) was proposed in \cite{mendel2014determining}. The Person FOU approach proceeds by taking an interval for each of the left and right endpoints of every word in the vocabulary by a single user, instead of a group of subjects. Then a uniform distribution is assumed to exist for both these data intervals and fifty random numbers are produced in each of the left and right intervals, denoted as $(L_1,…,L_{50})$ and $(R_1,…,R_{50})$, respectively. Following this, pairs of the form $(L_i,R_i )$,$i=1$ to $50$, are constructed by picking a value from each of the left and right intervals, so that now each data pair becomes an interval provided by $i^{th}$ (virtual) subject. 

The data intervals, collected in the case of perceptual computing or constructed in the case of Person FOU can be processed by the age-old Interval Approach (IA) \cite{liu2008encoding}. However, IA was later on identified to have some limitations. Therefore, the improved Enhanced Interval approach (EIA) \cite{wu2011enhanced} was proposed, which in turn was improved into the Hao-Mendel Approach (HMA) \cite{hao2015encoding}. The IA, EIA or HMA accomplishes the data processing using  the data part and the FS part. We will describe in detail each of the IA, EIA as well HMA, starting from IA. All the Equations used in the data and FS part of IA are given in Table \ref{tab:equationsIA}.

\subsubsection{IA: Data Part}
The data part is broadly divided into data pre-processing of the data intervals collected from a group of subjects (or constructed as in Person FOU) and statistics computation for the remaining data intervals from pre-processing. The pre-processing involves subjecting the data intervals to bad data processing, removal of the outliers, low confidence interval removal by tolerance limit processing, and reasonable-interval processing. We assume that the number of subjects are $n$ (there are 50 virtual subjects in Person FOU), and the endpoint data intervals for a word as $[a^{(i)}, b^{(i)}]$,$i=1,2,…,n$. 

\setlength\dashlinedash{0.5pt}
\setlength\dashlinegap{1.5pt}
\setlength\arrayrulewidth{0.3pt}

\newcommand\inlineeqno{\stepcounter{equation}\ (\theequation)}
\begin{sidewaystable}
    \centering
    \caption{Equations of IA}
    \begin{tabular}{l l}
    \hline
    \hline
         Name & Equation \\
         \hline\hline
         \textbf{Step 1:} Bad Data Processing & 
         
         $0 \leq a^{(i)} < b^{(i)} \leq 10, i=1,…,n ~ \inlineeqno\label{eq:baddata}$  \\
         \hdashline
         \textbf{Step 2:} Outlier processing & $\begin{rcases} a^{(i)} & \in [Q^a(0.25)-1.25IQR^a, Q^a(0.75)+1.5IQR^a]\\ b^{(i)} & \in [Q^b(0.25)-1.25IQR^b, Q^b(0.75)+1.5IQR^b]\\ L^{(i)} & \in [Q^L(0.25)-1.25IQR^L, Q^L(0.75)+1.5IQR^L]\end{rcases}\text{i=1,…,n'} ~\inlineeqno\label{eq:outlier}$\\
         \hdashline
         \textbf{Step 3:} Tolerance limit processing & $ \begin{rcases} a^{(i)} & \in [m^a-ks^a, m^a+ks^a]\\ b^{(i)} & \in [m^b-ks^b, m^b+ks^b]\\ L^{(i)} & \in [m^L-ks^L, m^L+ks^L]\end{rcases}\text{i=1,…,m'} ~\inlineeqno\label{eq:tolerancelimit}$\\
         \hdashline
         \textbf{Step 4:} Reasonable interval & $ a^{(i)} < \varepsilon^* < b^{(i)}, i=1,…m" ~\inlineeqno\label{eq:reasonableinterval1}$ \\
         processing & $\varepsilon^*=\frac{(m^b(s^a)^2 - m^a(s^b)^2) \pm s^as^b[(m^a-m^b)^2+2((s^a)^2-(s^b)^2)\ln{\frac{s^a}{s^b}}]^{\frac{1}{2}}}{(s^a)^2 - (s^b)^2} ~\label{eq:reasonableinterval2}\inlineeqno$\\
         \hdashline
         \textbf{Step 5:} Computing the statistics  & $S_i=(m_Y^{(i)},s_Y^{(i)}),i=1,…,m ~\label{eq:computingstatistics1}\inlineeqno$\\
         for surviving data intervals &$m_Y^{(i)}=\frac{(b^{(i)}+a^{(i)})}{2}, s_Y^{(i)}= \frac{(b^{(i)}-a^{(i)})}{\sqrt{12}}, i=1,…,m ~$\\
         \hdashline
        \textbf{Step 6:}  Computing the & $\label{eq:computinginteriormf}I~MF^1: m_{MF} = \frac{(b_{MF}+a_{MF})}{2}, s_{MF} = \frac{(b_{MF}-a_{MF})}{2\sqrt{6}} ~\label{eq:computingstatistics2}\inlineeqno$\\
         uncertainty measures for T1 & $L~MF^1:m_{MF} = \frac{(b_{MF}+2a_{MF})}{3}, s_{MF} = \left[\frac{1}{6} [(a_{MF}+b_{MF})^2+2a_{MF}^2]-m_{MF}^2\right]^{\frac{1}{2}} ~\label{eq:computinglsmf}\inlineeqno$\\
         FS models & $R~MF^1:m_{MF} = \frac{(b_{MF}+2a_{MF})}{3}, s_{MF} = {\left[\frac{1}{6}[{({a'}_{MF}{}+{b'}_{MF})}^{2}+{{2a'}_{MF}}^{2}]-{m'}_{MF}^{2}\right]}^{\frac{1}{2}} ~\label{eq:computingrsmf}\inlineeqno$\\
         & $a'_{MF}=M-b_{MF}, b'_{MF}= M-a_{MF} ~and ~m'_{MF}=M-m_{MF}$\\
         \hdashline
         \textbf{Step 7:} Computing general & $I~MF^1: a_{MF}^{(i)} = \frac{1}{2}\left[(a^{(i)}+b^{(i)})-\sqrt{2}(b^{(i)}-a^{(i)})\right], b_{MF}^{(i)} = \frac{1}{2}\left[(a^{(i)}+b^{(i)})+\sqrt{2}(b^{(i)}-a^{(i)})\right] ~\label{eq:computinggeneralformulainterior}$\\
          formulae for parameters of T1& $L~MF^1: a_{MF}^{(i)} = \frac{(a^{(i)}+b^{(i)})}{2}-\frac{(b^{(i)}-a^{(i)})}{\sqrt{6}}, b_{MF}^{(i)} = \frac{(a^{(i)}+b^{(i)})}{2}+\frac{\sqrt{6}(b^{(i)}-a^{(i)})}{3} ~\inlineeqno$\\
          FS models & $R~MF^1: a_{MF}^{(i)} = M- \frac{(a'^{(i)}+b'^{(i)})}{2}-\frac{\sqrt{6}(b'^{(i)}-a'^{(i)})}{3} b_{MF}^{(i)} = M - \frac{(a'^{(i)}+b'^{(i)})}{2}+\frac{(b'^{(i)}-a'^{(i)})}{\sqrt{6}} ~\label{eq:computinggeneralformularsmf}\inlineeqno$\\
         & $a'^{(i)}= M-b^{(i)}, b'^{(i)}= M-a^{(i)}$ \\
         \hdashline
         \textbf{Step 8:} Calculating the \\embedded T1 FSs & $(a^{(i)}, b^{(i)}) \rightarrow (a_{MF}^{(i)}, b_{MF}^{(i)}), i= 1,2,...,m ~\inlineeqno\label{eq:calculatingembedded}$\\
         \hdashline
         \textbf{Step 9:} Deleting the \\inadmissible T1 FSs & $a_{MF}^{(i)} \geq 0 ~and~ b_{MF}^{(i)} \leq 10, ~i=1,…,m ~\label{eq:deletinginadmissible}$\\
         \hdashline
         \textbf{Step 10:} Computing an IT2 FS & $\tilde{W} = \bigcup\limits_{i=1}^{m^*} W^{(i)} ~\inlineeqno\label{eq:computingit2}$\\
         \hdashline
         \textbf{Step 11:} Computing the & $\underline{a}_{MF} \equiv \min\limits_{i=1,…..,m^*}\{a_{MF}^{(i)}\}, \bar{a}_{MF} \equiv \max\limits_{i=1,…..,m^*}\{a_{MF}^{(i)}\}~\label{eq:computingfou}$\\
         mathematical model for FOU & $\underline{b}_{MF} \equiv \min\limits_{i=1,…..,m^*}\{b_{MF}^{(i)}\}, \bar{b}_{MF} \equiv \max\limits_{i=1,…..,m^*}\{b_{MF}^{(i)}\} $\\
         & $C_{MF}^{(i)} = \frac{a_{MF}^{(i)}+b_{MF}^{(i)}}{2}, \underline{C}_{MF} \equiv \min_{}\{C_{MF}^{(i)}\}, \bar{C}_{MF} \equiv \max_{}\{C_{MF}^{(i)}\}$\\
    \hline
    \hline
    \multicolumn{2}{l}{$^1$I MF = Interior Membership Function, L MF = Left Shoulder Membership Function,} \\
    \multicolumn{2}{l}{$^1$R MF = Right Shoulder Membership Function} \\
    \end{tabular}
    \label{tab:equationsIA}
\end{sidewaystable}

\paragraph*{\textbf{Step-1: Removing bad data}} Bad data is synonymous with the data intervals that either lie outside the assumed information scale of 0 to 10 or where the lower endpoint value of the interval is greater than (or equal to) the right endpoint value. All such data intervals are considered unsuitable for further processing and hence dropped from the set of useful data intervals by performing the comparisons given in (\ref{eq:baddata}). Hence, some of the data intervals may be dropped, thereby reducing the number of data intervals to $n'$, $n \geq n'$.

\paragraph*{\textbf{Step-2: Outliers removal}} This is the next step within data pre-processing. The outliers are identified in the $n'$ remaining data intervals through Box and whisker test. These computations are shown in (\ref{eq:outlier}), where $Q^a(0.25)$, $Q^b(0.25)$, and $Q^L(0.25)$ are the first quartiles for the interval left endpoints, right endpoints and lengths, respectively. $Q^a(0.75)$, $Q^b(0.75)$, and $Q^L(0.75)$ are the corresponding third quartiles, whereas the $IQR^a$, $IQR^b$, and $IQR^L$ are the corresponding words, interquartile ranges. The data intervals appearing as outliers in the test are filtered out, thereby reducing the number of data intervals to $m'$, $n' \geq m'$.

\paragraph*{\textbf{Step-3: Removing low confidence data intervals using tolerance limit calculation}} Tolerance factor $k$ provides $100(1-\gamma)\%$ confidence that an interval contains at least the proportion $(1-\alpha)\%$ of data values. The tolerance limit calculation is shown in (\ref{eq:tolerancelimit}), where $m^a$, $m^b$ and $m^L$ are the mean values for interval left endpoints, right endpoints and lengths, respectively. In contrast, $s^a$, $s^b$ and $s^L$ are the respective standard deviation values. Thus, out of the $m'$ surviving data intervals from the previous step, the ones not satisfying (\ref{eq:tolerancelimit}) are dropped from further processing, thereby giving rise to a reduced set of data intervals totalling $m"$, $m' \geq m"$.

\paragraph*{\textbf{Step-4: Reasonable interval processing}} For the $m''$ surviving data intervals from the previous step, computations are performed as given in (\ref{eq:reasonableinterval1}), using an optimal value of a threshold $\varepsilon^*$. In (\ref{eq:reasonableinterval2}), $m^a$ and $m^b$ are the mean of the interval left and right endpoints, respectively, whereas $s^a$ and $s^b$ are the corresponding standard deviation values, for surviving $m"$ data intervals of tolerance limit processing. The intervals satisfying (\ref{eq:reasonableinterval1}) have a high overlap with other intervals and hence are retained, and others are possibly dropped, thereby attracting a trimming in the numbers of data intervals to $m$, $m" \geq m$.

\paragraph*{\textbf{Step-5: Statistics computation for data intervals}} In this step, mean and standard deviation are computed for all the $m$ remaining data intervals from the previous step, assuming a probability distribution. The computations for this step are given in (\ref{eq:computingstatistics1}) and (\ref{eq:computingstatistics2}), assuming uniform probability distribution. 

\subsubsection{IA: FS Part}
The possible remaining $n$ data intervals obtained from the data part are processed using the FS part in various steps, as discussed now.

\paragraph*{\textbf{Step-6: Selecting T1 FS models}} In this step, the T1 MFs are classified into either left-shoulder, interior or right shoulder based on the mean and standard deviation values computed in (\ref{eq:computingstatistics1}) and (\ref{eq:computingstatistics2}). 

\paragraph*{\textbf{Step-7: Determining FS uncertainty measures}} After identifying the T1 MFs to belong to one of the three types viz., interior, left shoulder, and right shoulder, the two FS uncertainty measures: mean and standard deviation, are established for all the T1 MFs.

\paragraph*{\textbf{Step-8: Computing the uncertainty measures for T1 FS models}} Now, the mean and standard deviation FS uncertainty measures are computed for T1 FS models as shown in (\ref{eq:computinginteriormf})-(\ref{eq:computingrsmf}).  In these equations, $a_{MF}$ and $b_{MF}$ are the points on $x$-axis where the ends of T1 MF rest. 

\paragraph*{\textbf{Step-9: General formulae for T1 FS model parameters}} In this step, the FS uncertainty measure values of the T1 MFs (\ref{eq:computingstatistics1}) and (\ref{eq:computingstatistics2}) are equated to the corresponding values from (\ref{eq:computinginteriormf})-(\ref{eq:computingrsmf}). This is done in order to compute the general formulae for parameters defining the T1 FS models, which are given in (\ref{eq:computinggeneralformulainterior})-(\ref{eq:computinggeneralformularsmf}).

\paragraph*{\textbf{Step-10: Establishing the type of FOU}} Now, it is established that the FOU belongs to which one of the interior or shoulder FOUs by using the $t$-test on the mean and standard deviation values computed in (\ref{eq:computinggeneralformulainterior})-(\ref{eq:computinggeneralformularsmf}).

\paragraph*{\textbf{Step-11: Computing the embedded T1 FSs}} In this step, the T1 FSs, called the embedded T1 FSs, are obtained for the $m$ data intervals (after deciding on whether the FSs are interior or shoulder FOUs), by using the computations shown in (\ref{eq:calculatingembedded}). The embedded FSs are denoted as $W^{(i)}$,$i=1,…,m$\footnote{The $m$ number of data intervals were obtained at the end of Data part.}.

\paragraph*{\textbf{Step-12: Deleting the inadmissible T1 FSs}} This step involves the removal of inadmissible intervals, which fail to satisfy the (\ref{eq:deletinginadmissible}). This may be followed by reducing the number of data intervals from $m$ to $m^*$, $m \geq m^*$.

\paragraph*{\textbf{Step-13: Computing an IT2 FS word models}} Here, the wavy slice representation theorem comes into play for the computation of the IT2 FS word model using the embedded T1 FSs from the previous step. The computation for the IT2 FS word model is shown in (\ref{eq:computingit2}). 

\paragraph*{\textbf{Step-14: Computing the mathematical model for FOU}} In this final step, the mathematical model of the FOU viz., $FOU(\tilde{W})$ is computed as shown in (\ref{eq:computingfou}). Once this mathematical model is computed, the parameters for the Upper MF and the Lower MF are computed for all the IT2 FS models, which are denoted as $UMF(\tilde{W})$ and $LMF(\tilde{W})$), respectively.

\begin{figure}[h]
    \centering
    \includegraphics[height=\textheight, width=\textwidth, keepaspectratio=true]{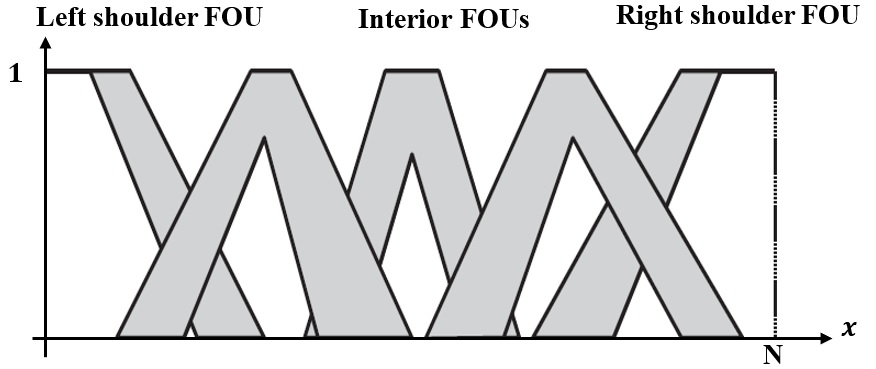}
    \caption{The IT2 FS word models obtained with IA}
    \label{fig:fousIA}
\end{figure}

In this manner, the IT2 FS word models are generated for every word of the vocabulary, using the data part and FS part. The IT2 FS word models belong to one of three categories viz., left shoulder, interior or right shoulder FOU as shown in Fig. \ref{fig:fousIA}. The obtained IT2 FS word models are defined using the nine point parameters: four for the UMF and five for the LMF, which are collectively called the FOU data. The FOU data, along with corresponding words, are stored in the form of a codebook.

\subsubsection{EIA \& HMA: Data Part}
EIA was proposed to improve the data processing in the data part of the IA, to ensure the better acquisition of the targeted data intervals. Hence, the EIA differs in the data pre-processing data part, though EIA also consists of the data part and the FS part. So, we will discuss the differences only here. Again, let's assume that our data collection set involves data intervals provided by $n$ subjects (there are 50 virtual subjects in Person FOU) and their data intervals are in the form $[a^{(i)},b^{(i)}]$, $i=1,2,…,n$. When these data intervals are subjected to the data pre-processing, we proceed as follows.

\renewcommand{\thefootnote}{\fnsymbol{footnote}}
\setcounter{footnote}{0}
\paragraph*{\textbf{Step-1\protect\footnote{replace the same step in IA}: Bad data processing}} In this step, the IA ensures that the endpoints of the interval lie within the scale of 0 to 10 as well as the left endpoint of the interval is less than the right one. The EIA imposes these constraints on the interval length also. These computations are shown in (\ref{eq:eiabaddata}). Thus, the intervals satisfying these computations are accepted, which may trim the number of data intervals to $n'$ from $n$.
    \begin{equation}
        \label{eq:eiabaddata}
        0 \leq a^{(i)} < b^{(i)} \leq 10, b^{(i)} - a^{(i)} < 10, i=1,…,n
    \end{equation}

\paragraph*{\textbf{Step-2\protect\footnotemark[\value{footnote}]: Outlier processing}} The EIA initially identifies the outliers within the surviving $n'$ interval endpoints using Box and whisker test, thereby leading to a possible reduced set of data intervals totalling $n"$, $n' > n"$. The computations are given as:

\begin{equation}
    \label{eq:eiaoutlier1}
    \begin{rcases}
        & a^{(i)} \in [Q^a(0.25)-1.25\text{IQR}^a, Q^a(0.75)+1.5\text{IQR}^a]\\ 
        & b^{(i)} \in [Q^b(0.25)-1.25\text{IQR}^b, Q^b(0.75)+1.5\text{IQR}^b]
    \end{rcases} \text{i=1,…,n'}\\
\end{equation}

Here, $Q^a(0.25)$, $Q^a(0.75)$ and $IQR^a$ are respectively the first quartiles, third quartiles and interquartile range for the interval left endpoints. The $Q^b(0.25)$, $Q^b(0.75)$ and $IQR^b$ are the corresponding values for the interval right endpoints. 

Following this, the outliers are identified for the interval lengths of the $n"$ surviving data intervals, using again the Box and whisker test. These computations are given as:
	
\begin{equation}
    \label{eq:eiaoutlier2}
        L^{(i)} \in [Q^L(0.25)-1.25\text{IQR}^L, Q^L(0.75)+1.5\text{IQR}^L], i = 1,2,..., n"
\end{equation}

Here, $Q^L(0.25)$, $Q^L(0.75)$ and $IQR^L$ are, respectively, the first quartile, third quartile and interquartile range. Thus, after performing an additional Box and Whisker test sequentially on the interval endpoints and on the interval lengths, there is a possibility of trimming the number of data intervals to $m'$, $m' \leq m'$. 

\paragraph*{\textbf{Step-3\protect\footnotemark[\value{footnote}]: Tolerance limit processing}} In EIA, the tolerance limit processing also enables us to identify high confidence intervals. Likewise the previous step, the tolerance limit processing is applied on the interval endpoints, using a confidence factor $k$, shown as:

\begin{equation}
    \label{eq:eiatolerance1}
    \begin{rcases} 
	    a^{(i)} & \in [m^a-ks^a, m^a+ks^a]\\ 
	    b^{(i)} & \in [m^b-ks^b, m^b+ks^b]\\ 
	 \end{rcases}\text{i=1,…,m'}
\end{equation}

Here, $m^a$ and $s^a$ are the mean and standard deviation of the interval left endpoints. $m^b$ and $s^b$ are the corresponding  values for interval right endpoints. The computations in (\ref{eq:eiatolerance1}) may lead to a reduction in the number of data intervals from $m'$ to $m^+$. Following this, the computations are performed on interval lengths of the surviving $m^+$ data intervals, using a confidence factor $k'$ as:

\begin{equation}
    \label{eq:eiatolerance2}
	    L^{(i)} \in [m^L-k's^L, m^L+k's^L], i=1,…,m^+
\end{equation}
where $m^L$ is the mean of the interval length and $s^L$ is the standard deviation. $k'$ in (\ref{eq:eiatolerance2}) is given as:

\begin{equation}
    \label{eq:eiatolerance3}
	    k'= min(k_1, k_2, k_3 )
\end{equation}

The value of $k_1$ enables that one can state with $95\%$ confidence that the interval $[m^L-k_1 s^L, m^L+k_1 s^L]$ contains at least the proportion $95\%$ of data values. The values $k_2$ and $k_3$ are given as:

\begin{equation}
    \label{eq:eiatolerance4}
	    k_2 = \frac{m^L}{s^L}, k_3 = \frac{(10-m^L)}{s^L}
\end{equation}

The computations in (\ref{eq:eiatolerance4}) are performed to retain only the not too small or not too large intervals. Hence, at the end of the tolerance limit processing step, the number of data intervals may be reduced to $m"$.

\paragraph*{\textbf{Step-4\protect\footnotemark[\value{footnote}]: Reasonable interval processing}} Reasonable interval processing accepts those intervals that have a high amount of overlap with other intervals. These intervals are identified using the computations shown in (\ref{eq:eiareasonable}):

\begin{equation}
    \label{eq:eiareasonable}
        a^{(i)} < \varepsilon^* < b^{(i)}, 2m^a- \varepsilon^* \leq a^{(i)}, b^{(i)} \leq 2m^b- \varepsilon^*, ~i=1,…m"
\end{equation}

here $m^a$ and $s^a$ are the mean and standard deviation, respectively, of the left endpoints. The $m^b$ and $s^b$ are the corresponding values for the right endpoints. Further, $\varepsilon^*$ is the optimal value of the threshold, which is computed as shown in (\ref{eq:reasonableinterval2}). Finally, at the end of reasonable interval processing, the set of data intervals may be trimmed to a length of $m$. 

\subsubsection{EIA: FS part}
The EIA FS part performs all the computations in the same nine steps as that in IA, except in the last step viz., calculating the mathematical model of FOU, where the LMF height of interior FOUs is calculated in such a manner that it avoids completely filled and flat FOUs. 

Hence, EIA also generates the IT2 FS word models for all the vocabulary words and stores them in a codebook. It is mentioned here that EIA FOUs are also either interior or shoulder, as shown in Fig. \ref{fig:fousIA}.

\subsubsection{HMA: FS part}
HMA improves the FS part IA/EIA in order to extract more information from the data intervals. Upon receiving the $m$ surviving data intervals from the data part, they are processed for overlap computation in the FS part. Then the IT2 FS word models are then computed using a smaller set of remaining data intervals that do not contain the overlap. Let's discuss the working of the various steps in the FS part in detail. It is mentioned here that only the steps in the FS part of HMA that differ from EIA are discussed here.

\paragraph*{\textbf{Step-6\protect\footnotemark[\value{footnote}]: Establishing the nature of FOU}} To establish whether a data interval may be associated to an interior or a shoulder FOU, one-sided tolerance limits are calculated for the $m$ remaining data intervals of the data part. This is given in (\ref{eq:hmaestablishingfou}) as:

\begin{equation}
    \label{eq:hmaestablishingfou}
    \underline{a} = \hat{m}_a - k(m)\hat{s}_a, ~\bar{b} = \hat{m}_b - k(m)\hat{s}_b
\end{equation}

where $\hat{m}_a$ and $\hat{s}_a$ are respectively the sample mean and standard deviation for $m$ data intervals’ left endpoints. The $\hat{m}_b$ and $\hat{s}_b$ are the corresponding values for the right endpoints. $k(m)$ is the one-sided tolerance factor (and a function of $m$). Given the value of the one-sided tolerance factor, it can be stated with $95\%$ confidence that the given limit contains greater than or equal to $95\%$ data. 

After computing these one sided tolerance limits, the word model ($W$) is classified into the interior or one of the shoulder FOUs based on these comparisons:

\begin{equation}
    \label{eq:hmafouclassification}
    \begin{cases}
    if \underline{a} \leq 0: Left Shoulder FOU\\
    if \bar{b} \geq 10: Right Shoulder FOU \\
    Otherwise: Interior FOU
    \end{cases}
\end{equation}

\paragraph*{\textbf{Step-7\protect\footnotemark[\value{footnote}]: Computing the interval overlap}} Here, the computations for the overlap intervals $[o_a,o_b]$ are performed for the interior as well as the shoulder FOUs as:

\begin{equation}
    \label{eq:hmaoverlap}
    [o_a,o_b ] = 
    \begin{cases}
    [0, \min\limits_{i}b^{(i)}], i= 1,...,m: Left Shoulder FOU\\
    [\max\limits_{i}a^{(i)}, 10], i= 1,...,m: Right Shoulder FOU \\
    [\max\limits_{i}a^{(i)}, \min\limits_{i}b^{(i)}], i= 1,...,m: Interior FOU
    \end{cases}
\end{equation}

\paragraph*{\textbf{Step-8\protect\footnotemark[\value{footnote}]: Removing the interval overlap}} The purpose of calculating the overlap intervals $[o_a,o_b]$ in the previous step was to remove them from all the $m$ data intervals to arrive at a smaller sets of data intervals for the interior and the shoulder FOUs by performing the computations as:

\begin{equation}
\label{eq:hmaremovingoverlap}
\begin{split}
    &[o_b=\min\limits_{i}⁡b{(i)}, ⁡b{(i)} ], ~i=1,…,m:  {Left shoulder FOU}\\
    &[a^{(i)}, o_a=\max\limits_{i}a^{(i)}], ~i=1,…,m:  {Right shoulder FOU} \\
    &[a^{(i)}, o_a=\max\limits_{i}a^{(i)}] ~\& ~ [o_b=\min\limits_{i}⁡b{(i)}, ⁡b{(i)}]{,i=1,...m}: {Interior FOU} \\
    \end{split}
\end{equation}

\begin{figure}[h]
    \centering
    \includegraphics[height=\textheight, width=\textwidth, keepaspectratio=true]{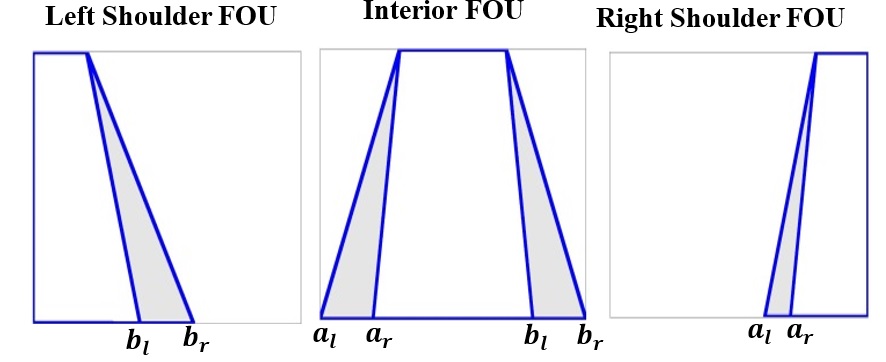}
    \caption{FOU plots obtained with HMA}
    \label{fig:fousHMA}
\end{figure}

\paragraph*{\textbf{Step-9\protect\footnotemark[\value{footnote}]: Mapping the data intervals into FOU parameters}} Once the smaller set of intervals is obtained using the computations in (\ref{eq:hmaremovingoverlap}), the quantities in the equation are mapped into the FOU parameters of the interior or shoulder FOUs. For left shoulder FOU the parameters $[b_l,b_r]$ are mapped to the respective interval values from (\ref{eq:hmaoverlap}). The same action follows for the parameters $[a_l,a_r]$ of the right shoulder FOU. The interior FOU mapping is slightly different, and looks like: $[a^{(i)}, o_a=\max\limits_{i}a^{(i)}]$ to $[a_l,a_r]$, and $[o_b=\min\limits_{i} b^{(i)}, b^{(i)}]$ to $[b_l,b_r]$. 

Based on these parameter values, the obtained FOU plots with the HMA look similar to the ones shown in Fig. \ref{fig:fousHMA}. By visual inspection, one can make out the difference between the IA/ EIA FOU plots shown in Fig. \ref{fig:fousIA} and HMA FOU plots shown in Fig. \ref{fig:fousHMA}. In the IA/ EIA IT2 FS word models, the UMF has a height of 1. The LMF may have a height of 1 for the shoulder IT2 FS word models and generally not for the interior ones. The HMA IT2 FS word models, on the other hand, plot a height of 1 for both the UMF and the LMF. 

\subsection{CWW Engine}\label{subsec:perccwwengine}
CWW engine performs the task of aggregating the user feedback in the form of IT2 FS word models for the codebook words. Whenever multiple stakeholders provide their inputs, at the encoder, in the form of words, these words are converted to IT2 FS word models as outlined above in Section \ref{subsec:percencoder}. Their nine point data about the IT2 FS word models are extracted from the codebook and given as input to the CWW engine. CWW engine can use different types of aggregation operators like interval weighted average (IWA), fuzzy weighted average (FWA), linguistic weighted average (LWA), etc. The differentiating factor for the use of these operators is the nature of the data to be aggregated. IWA is used when the data and the weights are no more than the intervals. FWA is used when at least one piece of information from the data or the weights are T1 FSs but not IT2 FSs, whereas the LWA is used when at least one piece of information from the data or the weights are IT2 FSs. 

The computations for aggregating the IT2 FS word models for the words and their associated weights using the LWA are accomplished as:

\begin{equation}
    \label{eq:perccwwenginelwa}
    \tilde{Y}_{LWA} = \frac{\sum_{i=1}^{n}\tilde{X}_i\tilde{W}_i}{\sum_{i=1}^{n}\tilde{W}_i}
\end{equation}

Here $\tilde{X}_i$ is the IT2 FS models of the words to be aggregated and $\tilde{W}_i$ are those of the corresponding weights. 

\subsection{Decoder}\label{subsec:percdecoder}
The output of the CWW engine viz., $\tilde{Y}_{LWA}$, is also an IT2 FS and is generally not an exact match to any of the IT2 FS word models in the codebook. Hence, the decoder section performs the task of generating linguistic recommendations by performing various computations. The linguistic recommendation or the ‘word’ is generated using Jaccard’s similarity measure. 

The decoder is also capable of generating ‘ranking’ and ‘class’ as recommendations. The most commonly used ranking method is the centroid ranking, where the centroid of an IT2 FS is given by the union of the centroids of all its embedded T1 FSs, $c(A_e)$, as shown in (\ref{eq:perdecoderunionembedded}):

\begin{equation}
    \label{eq:perdecoderunionembedded}
    C_{\tilde{A}(x)} =  \bigcup\limits_{\forall A_e}^{} c(A_e) \equiv [c_l, c_r]
\end{equation}

The values of $c_l$  and $c_r$ in (\ref{eq:perdecoderunionembedded}) are given as:

\begin{equation}
    \label{eq:percdecodercentroid1}
    c_l = \frac{\sum_{i=1}^{L}x_i\bar{\mu}_{\tilde{A}}(x_i) + \sum_{i=L+1}^{N}x_i\underline{\mu}_{\tilde{A}}(x_i)}{\sum_{i=1}^{L}\bar{\mu}_{\tilde{A}}(x_i) + \sum_{i=L+1}^{N}\underline{\mu}_{\tilde{A}}(x_i)}
\end{equation}

\begin{equation}
    \label{eq:percdecodercentroid2}
    c_r = \frac{\sum_{i=1}^{R}x_i\underline{\mu}_{\tilde{A}}(x_i) + \sum_{i=R+1}^{N}x_i\bar{\mu}_{\tilde{A}}(x_i)}{\sum_{i=1}^{R}\underline{\mu}_{\tilde{A}}(x_i) + \sum_{i=R+1}^{N}\bar{\mu}_{\tilde{A}}(x_i)}
\end{equation}

\begin{figure}[h]
    \centering
    \includegraphics[height=\textheight, width=\textwidth, keepaspectratio=true]{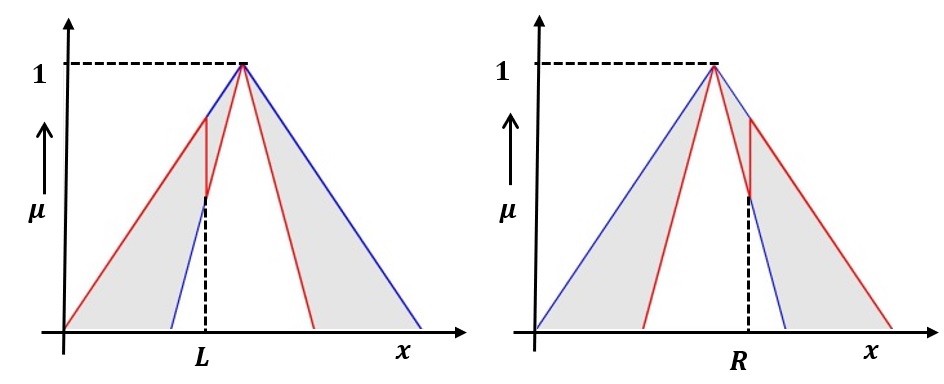}
    \caption{Switch Points for Centroid Calculation in IT2 FS}
    \label{fig:switchpoints}
\end{figure}

In (\ref{eq:percdecodercentroid1})-(\ref{eq:percdecodercentroid2}), $L$ and $R$ are called switch points at which the aggregation switches between the UMF and LMF. They can be computed using EIASC algorithm, Karnik Mendel (KM) algorithm and Enhanced Karnik Mendel (EKM) algorithm (Fig. \ref{fig:switchpoints}).

The mean, $c(\tilde{A})$, of $c_l$ and $c_r$ is used in the centroid ranking method and computed as:

\begin{equation}
    \label{eq:perdecodercentroidmean}
    c(\tilde{A})=\frac{(c_l+c_r)}{2}
\end{equation}

\section{CWW based on T1 FS extensions}\label{sec:l-4cwwmethodologies}
In this section, we discuss the details of the CWW methodologies that model the semantics of the linguistic terms using the T2 FSs, and hence called the CWW methodologies based on the extension of T1 FSs. These are: LFSCM and the GFSCM. In the LFSCM, the secondary MF of the shoulder FOUs is a linear function. On the other hand, in GFSCM, the MF of the interior, as well as shoulder FOUs, need not necessarily be a linear function. 

\subsection{LFSCM}\label{subsec:lfscm}
\begin{figure}[h]
    \centering
    \includegraphics[height=\textheight, width=\textwidth, keepaspectratio=true]{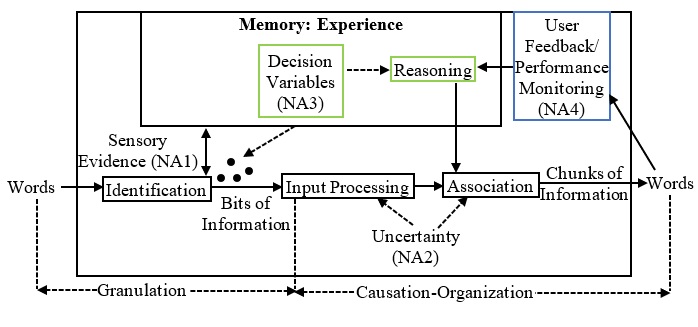}
    \caption{Architecture for LFSCM \cite{bilgin2015ambient}}
    \label{fig:lgt2cww}
\end{figure}

LFSCM models the word semantics using the T2 FSs, where the secondary MF is a linear function, and hence these FSs are called the LGT2 FSs \cite{bilgin2015ambient}. The LFSCM uses a CWW architecture, as shown in Fig. \ref{fig:lgt2cww}. The LFSCM consists of mainly two parts viz., granulation and causation-organization (with each part consisting of several other interacting components). Granulation works on the principle of decomposing a whole unit into parts, whereas causation associates the causes with effects. The organization works opposite to the granulation by integrating parts into a whole. Inside the LFSCM, a sub-component exists commonly referred to as the `neural architecture for perceptual decision-making’, which contains four modules viz., Sensory Evidence or NA1 (responsible for accumulation and comparison of sensory evidence), Uncertainty or NA2 (has the task of perceptual uncertainty detection), Decision Variables or NA3 (decision variables are represented) and User Feedback/ Performance Monitoring or NA4 (monitors the performance by detecting errors and adjustment of decision strategies). Further, the system requires Memory for data accumulation and hence it is also one of the components. In the terminology of the neural architecture, it is commonly referred to as the Experience.\\ 

\subsubsection{Step-1: Granulation}
The operation of the LFSCM is triggered by the problem specific input `words', to the granulation segment. Here, these words are mapped into their numerical representations based on the sensory evidence. The sensory evidence acts as a solution descriptor relates to decision variables in human reasoning.\\

\subsubsection{Step-2: Causation-Organization}
In the next segment, viz., causation-organization, these numerical representations are associated with the appropriate fuzzy IF-THEN rules (fuzzy IF-THEN rules deal with the uncertainty of the human reasoning). The output of these IF-THEN rules is aggregated and converted back to the linguistic form, for communication to the user. The words are displayed to the user through a Graphical User Interface (GUI). The user is asked to evaluate the output of the system and provide his/ her feedback. Thus, based on the user feedback, the IF-THEN rules can be modified to incorporate the user preferences. Further, it is pertinent to mention that for the first time, the system output is generated based on the default rule configuration. Subsequent user interaction enables the populating of IF-THEN rule base with user preference based rules, and hence more user acceptable output words can be obtained from the system.\\
\subsection{GFSCM}\label{subsec:gfscm}
A GFSCM using various types of fuzzy automata was proposed in \cite{jiang2018general}, of which we discuss here the one based on universally GT2 fuzzy finite automata. According to the definition given in \cite{jiang2018general}, the universally GT2 fuzzy finite automata for CWW is a quin tuple given as: $M=(Q,\sum',\delta,q_0,F)$. Here, the meanings of each of the quantities in the quin tuple are given as:
\begin{itemize}
    \item $Q=\{q_0,q_1,…,q_N\}$ is the finite set of states.
    \item The $\sum'$ is the finite set of symbols, so-called the underlying input alphabet. It is mentioned that $\sum'$ is the type-2 fuzzy subset of $GT2F(\sum)$, where $\sum$ is the finite input alphabet.
    \item $\delta$ is the type-2 fuzzy transition function defined as: $\delta : Q \times \sum' \rightarrow GT2F(Q)$. Thus, for any $q_i \in Q$ and $a \in \sum'$, $\delta (q_i,a)$ may be considered as a possibility distribution of the states that the automata in state $q_i$ can enter given the input alphabet $a$. This can be written more generally as $\delta (q_i,a)(q_j)$, $q_j$ being the state to enter from $q_i$ with the given input alphabet $a$.
    \item $q_0 \in Q$ is the initial state.
    \item $F$ is the GT2 fuzzy subset of $Q$, called the GT2 FS of final states. More generally, $F(q_i), \forall q_i \in Q$, denotes the degree to which $q_i$ is the final state.
\end{itemize}

In particular, when $\sum'= GT2F(\sum)$, then the universal GT2 automata is the fuzzy automata for CWW (for all words).

\section{Discussions}\label{sec:discussions}
We now discuss some important views related to the work presented here.

A remarkable work \cite{gupta2021enhanced}, has reasoned the supremacy of AEPCM and IFSCM over the EPCM as well as RSCM over the SMCM. Further, there has been a number of works in the literature where the IT2 FS based Perceptual Computing has been shown to possess much better performance than the EPCM, SMCM and 2TPCM. The details can be found here \cite{BibEntry2021Jul}. However, no comparison has been made so far between perceptual computing and AEPCM, IFSCM and RSCM. This remains an area worth exploring.

In \cite{bilgin2015ambient}, the LFSCM is shown to give almost $50\%$ better performance than the perceptual computing. Thus, it is safe to conclude that as we use the higher level FSs to model the `word' semantics, the model accuracy increases. However, all this comes at the expense of a higher computational cost.

It is quite evident that a lot of research work has been done in Perceptual computing. However, there has not been much research on the other CWW methodologies. Further, a recent novel work \cite{bustince2015historical}, gives a thorough literature review on various types of FSs. Out of these, there are various types of FSs on which CWW methodologies have not been proposed so far. Interested researchers can work on this line of research.

The CWW methodology perceptual computing \cite{mendel2010perceptual} models the word semantics using the IT2 FSs, the words come from human perception. However, the basis for perception \cite{loomis1986tactual, bruce2003visual} is information acquired through the senses as well as making something useful out of it. The process of how human beings perceive through the senses and acquire information has a lot of complex psychological processes acting underlying it. It is in itself a very broad research area. On the contrary, in all these works on perceptual computing, the basis for information acquisition is the data collected about the location of the endpoints of a word, on a scale of 0 to 10, which a human subject thinks. There is no such process involved in the perceiving of any information through the senses. It is just how a human subject has an opinion about the location of word endpoints. So, in our opinion, the word ``perception" is slightly misleading.

A research article that attempted to put forth the idea of achieving the CWW using Machine learning, was \cite{becker2001computing}. In this article, the author used the scenario of medical diagnostics as a bedrock and highlighted the use of linguistic information for expressing a physician's knowledge. The author uses the concept of fuzzy random variable based probability theory to model the linguistic semantics. Also, in \cite{kahraman2010renewable}, the authors used the concept of Choquet integral for achieving the CWW. However, in \cite{martinez20152, marti2012overview}, it has been advocated that, more often than not, the uncertainty in real-life observations is non-probabilistic in nature. It stems from the semantic uncertainty of linguistic information. Hence, the CWW methodologies \cite{becker2001computing, kahraman2010renewable} are not close to human cognition in processing linguistic information.

\textcolor{blue}{It has been stated in \cite{mendel2012challenges} that CWW should make use of fuzzy logic. FSs do a great job at modeling the intra and inter uncertainty in the semantics of linguistic terms in a manner close to human beings. However, they also suffer from some shortcomings. According to us, major amongst those is the membership function generation for modeling the semantics of linguistic terms in CWW. The T1 FSs based CWW approaches divide the information representation scale uniformly amongst the T1 fuzzy MFs and the ordinal term sets based CWW approaches assume the complete information to be modeled in terms of ordered locations of linguistic terms on the information representation scale. The 2TPCM uses a combination of both T1 MFs and the ordinal terms. Hence, all these three categories of CWW approaches have a reduced capability to model word semantics compared to IT2 FSs based perceptual computing of the CWW approaches based on T1 FS extensions. The IT2 FSs based perceptual computing, on the other hand, generate MFs by collecting data from a group of people. This has its limitations, like being time consuming, etc. and T1 extension based CWW approaches are computationally intensive. Therefore, to overcome the limitation of IT2 FS based CWW approach, work is being done in this direction \cite{gupta2022enhancedwmit2}. The challenge is to automate the MF generation methodology while modeling the word semantics in the best possible way.}

\textcolor{blue}{Recently there has been a surge in the area of symbolic learning \cite{diaz2022explainable}. It represents concepts using symbols and then relationships are defined amongst them. Further, they are increasingly being used in conjunction with sub-symbolic systems for various purposes. CWW is often considered synonymous to the symbolic learning. However, there are some subtle differences. CWW attempts to model the semantic uncertainty of the linguistic terms so as to bring it close to the human cognitive process. This is required so that if a computing system be developed using the CWW principles, it should be able to process the linguistic information in a manner similar to the human beings. To ensure this, it is paramount that the system handles the semantics uncertainty in a manner as close as possible to the human cognition. Further, as can be seen from Fig. \ref{fig:yagergen}, there is no learning involved in any CWW methodology. Its emphasis lies on mapping the linguistic information to numeric (step 1: translation) by using an instrument that can capture and model the linguistic uncertainty in the best possible way. This is followed by aggregation (step 2: manipulation) and finally generating linguistic recommendation back (step 3: retranslation).}

\section{Conclusions and Future Scope}\label{sec:conclusions}
Computing with Words (CWW) was proposed by Prof. Zadeh, as a novel approach that aims to impart computing machines with the capability to process linguistic information like human beings. There have been numerous notable works in the CWW, both application-based and theoretical. It has been successfully applied in various application areas like risk assessment, decision analysis, decision support systems, etc. 

On the theoretical side, an important line of work has been the development of various CWW methodologies, which has resulted in various diverse research works. These include research works on the EPCM, AEPCM, IFSCM, SMCM, RSCM, 2TPCM, Perceptual computing, LFSCM and GFSCM. However, to the best of our knowledge, the literature on these methodologies is mostly scattered and does not give an interested researcher a comprehensive overview of the notion and utility of these methodologies. Hence, we have addressed this issue in this work and given the readers a succinct but wide survey of the CWW methodologies. 

We have also highlighted the future research directions to build upon for the interested and motivated researchers. These include the performance comparisons between the various CWP methodologies (wherever non-existent), dealing with the trade-off between methodology accuracy and computational cost, and a direction for developing more CWP methodologies based on different types of FSs.

\appendix
\section{Basics of IFS}\label{sec:basicsofifs}
\begin{figure}[h]
    \centering
    \includegraphics[height=0.7\textheight, width=0.7\textwidth, keepaspectratio=true]{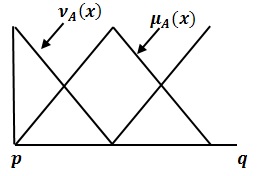}
    \caption{IFS: Representation}
    \label{fig:mem-nonmem-ifs}
\end{figure}

We give here a brief introduction to the IFSs \cite{intanssov1986intuitionistic, atanassov1988review, atanassov1989more, atanassov1999intuitionistic}. IFSs generalize the concept of T1 FSs. The T1 FS represents the uncertainty about each set element through its membership value. The IFS extend the notion of this uncertainty by an added quantity to each set element viz., the degree of non-membership. Mathematically, an IFS (A) is defined on a universe X as:

\begin{equation}
    \label{eq:ifs}
    A=\{(x,\mu_A(x),\nu_A(x) \mid x \in X\}
\end{equation}

where $\mu_A(x)$ and $\nu_A(x)$ is the degree of membership and degree of non-membership, respectively (Graphically are shown in Fig. \ref{fig:mem-nonmem-ifs}), both satisfying the condition: $0 \leq \mu_A(x)$, $\nu_A(x) \leq 1$, and $\mu_A(x)+\nu_A(x) \leq1$. An IFS differs from a T1 FS as in the latter, $\nu_A(x)=1-\mu_A(x)$ and law of excluded middle hold.
\\

\section{Indiscernibility of Rough Sets}\label{sec:indiscernibilityroughsets}
Indiscernibility is an important property of rough sets and forms a basis of various operations. To illustrate the concept, consider a system for modeling the information $I$. Let's assume that there is a universe of discourse $U$ and a non-empty finite attribute set $A$ defined on it. Let's say an attribute $a \in A$ exists and a set $V_a$ is defined, such that $a$ can take values from $V_a$. Then $I$ is a mapping defined in the following manner: $I:U \rightarrow V_a$. 

Let's say there is a mapping mechanism to assign a value $a(x)$ from $V_a$ to each attribute $a$ and object $x$ in the universe $U$. Then, $P$-indiscernibility, $IND(P)$, of a rough set defined mathematically as:

\begin{equation}
    \label{eq:roughsetsindescernability}
    IND(P)=\{(x,y) \in U^2 \mid \forall a \in P, a(x)=a(y)\}
\end{equation}

where $P \subseteq A$, forms an associated equivalence relation. The $U$ can be divided into a set of equivalence classes $IND(P)$ denoted by $U/IND(P)$ or $U/P$. 

\bibliographystyle{elsarticle-num-names} 
\bibliography{manuscript.bib}

\end{document}